\documentclass[journal]{IEEEtran}
\usepackage{verbatim}
\usepackage{multirow}
\usepackage{epsfig}
\usepackage{graphicx}
\usepackage{amsmath}
\usepackage{amssymb}
\usepackage{subfigure}
\usepackage{amssymb}
\usepackage{mathrsfs}
\usepackage{color}
\usepackage{float}
\usepackage{caption}
\usepackage{titlesec}

\usepackage[pagebackref=true,breaklinks=true,letterpaper=true,colorlinks,citecolor=citecolor,bookmarks=false]{hyperref}
\definecolor{citecolor}{RGB}{34,139,34}

\newcommand{\yx}[1]{\textcolor{black}{#1}}
\ifCLASSOPTIONcompsoc
  \usepackage[nocompress]{cite}
\else
  \usepackage{cite}
\fi

\newcommand{\ie}{\emph{i.e. }}
\newcommand{\etal}{\textit{et al.}}
\newcommand{\mycaption}{\caption*{\footnotesize}}

\ifCLASSINFOpdf
\else
\fi
\begin{document}
%
\title{ReGO: Reference-Guided Outpainting for \\ Scenery Image}
%
%
%

\author{Yaxiong Wang,
        Yunchao Wei, Xueming Qian, Li Zhu and Yi Yang
\IEEEcompsocitemizethanks{\IEEEcompsocthanksitem Y. Wang is with the School of Software Engineering, Xi'an Jiaotong University, Xi'an, 710049, China. \protect
E-mail: wangyx15@stu.xjtu.edu.cn.
\IEEEcompsocthanksitem Y. Wei is with Institute of Information Science, Beijing Jiaotong University, Beijing, 100000, China.\protect
E-mail: wychao1987@gmail.com.
\IEEEcompsocthanksitem X. Qian is with the Key Laboratory for Intelligent Networks and Network Security, Ministry of Education, Xi’an Jiaotong University, Xi’an 710049, China, also with the SMILES Laboratory, Xi’an Jiaotong University, Xi’an 710049,China, and also with Zhibian Technology Co. Ltd., Taizhou 317000, China.\protect
E-mail:qianxm@mail.xjtu.edu.cn.
\IEEEcompsocthanksitem L. Zhu is with the School of Software, Xi’an Jiaotong University, Xi’an 710049, China. \protect
E-mail: zhuli@mail.xjtu.edu.cn.
\IEEEcompsocthanksitem Yi Yang is School of Computer Science and Technology, Zhejiang University, Hangzhou, 310000, China. \protect
E-mail:yee.i.yang@gmail.com.}}

%
%

\markboth{Journal of \LaTeX\ Class Files,~Vol.~14, No.~8, August~2015}%
{Shell \MakeLowercase{\textit{et al.}}: Bare Demo of IEEEtran.cls for IEEE Journals}
%



\maketitle

\begin{abstract}
We aim to tackle the challenging yet practical scenery image outpainting task in this work. Recently, generative adversarial learning has significantly advanced the image outpainting by producing semantic consistent content for the given image. However, the existing methods always suffer from the blurry texture and the artifacts of the generative part, making the overall outpainting results lack authenticity. To overcome the weakness, this work investigates a principle way to synthesize texture-rich results by borrowing pixels from its neighbors (\ie, reference images), named \textbf{Re}ference-\textbf{G}uided \textbf{O}utpainting (ReGO). Particularly, the ReGO designs an Adaptive Content Selection (ACS) module to transfer the pixel of reference images for texture compensating of the target one. To prevent the style of the generated part from being affected by the reference images, a style ranking loss is further proposed to augment the ReGO to synthesize style-consistent results. Extensive experiments on two sceneary benchmarks, NS6K~\cite{yangzx}, NS8K~\cite{wang} and SUN Attribute~\cite{sunattr}, well demonstrate the effectiveness of our ReGO. Our code is available at  https://github.com/wangyxxjtu/ReGO-Pytorch.
\end{abstract}


%
\IEEEpeerreviewmaketitle

\section{Introduction}
%
%
%
%
\label{intro}
Given an input image, image outpainting aims at generating plausible visual content outside the image boundary. 
Traditional approaches~\cite{T_inpaint1,T_inpaint2,T_inpaint3,T_inpaint4} employ a simple searching and stitching pipeline, the extrapolation is achieved by stitching the picked image patches to the input image. However, these solutions are too inflexible to meet the practical requirements. Recently, inspired by the success of the Generative Adversarial Networks (GANs) \cite{GAN,wgan}, researchers are making efforts to synthesize the unseen content for the input image by adversarial learning~\cite{SRN,bds,yangzx,condition_progress,sketchGAN}. For example, Yang \etal~\cite{yangzx} propose a framework that could recurrently predict new content for the given image patch. In \cite{bds}, Teterwak~\etal translate the input image to a larger picture with new content beyond the boundary.  Wang \etal~\cite{wang} take the image outpainting one step forward by introducing the sketch clues to control the synthesis procedure. 

\begin{figure}[t]
\begin{center}
\subfigure[Input]{
    \begin{minipage}[t]{0.46\linewidth}
        \centering
        \includegraphics[width=1.64in]{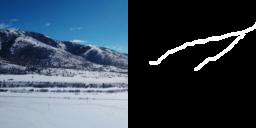}\\
        \label{fig1:a}
        \vspace{-0.4cm}
    \end{minipage}
    }
\subfigure[Yang \etal~\cite{yangzx}]{
    \begin{minipage}[t]{0.46\linewidth}
        \centering
        \includegraphics[width=1.64in]{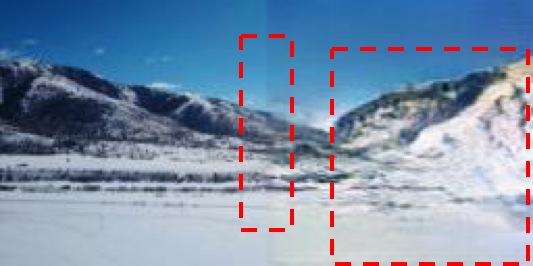}
        \label{fig1:b}
        \vspace{-0.65cm}
    \end{minipage}
    }
    \\
    \vspace{-0.2cm}
\subfigure[Teterwak \etal~\cite{bds}]{
    \begin{minipage}[t]{0.46\linewidth}
        \centering
       \includegraphics[width=1.64in]{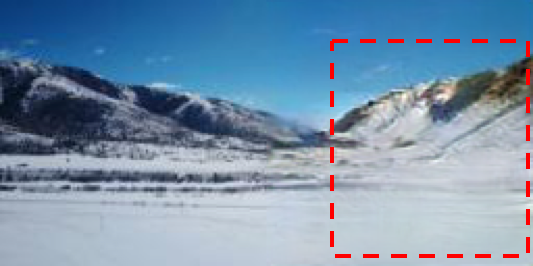}
        \label{fig1:c}
        \vspace{-0.2cm}
    \end{minipage}
    }
\subfigure[Wang \etal~\cite{wang}]{
    \begin{minipage}[t]{0.46\linewidth}
        \centering
        \includegraphics[width=1.64in]{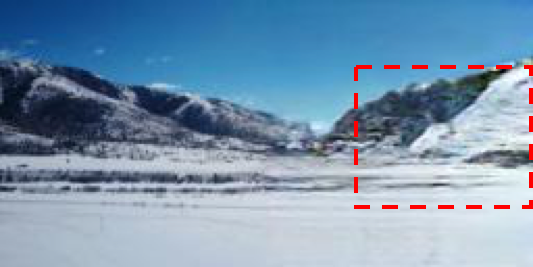}
        \label{fig1:d}
        \vspace{-0.2cm}
    \end{minipage}
    }
    \\
    \vspace{-0.2cm}
\subfigure[Ours]{
    \begin{minipage}[t]{0.46\linewidth}
        \centering
        \includegraphics[width=1.64in]{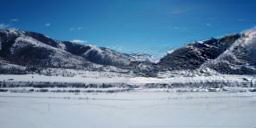}
        \label{fig1:e}
        \vspace{-0.2cm}
    \end{minipage}
    }
\subfigure[original image]{
    \begin{minipage}[t]{0.46\linewidth}
        \centering
         \includegraphics[width=1.64in]{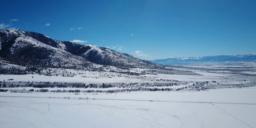}
        \label{fig1:f}
        \vspace{-0.2cm}
    \end{minipage}
    }
\vspace{-0.2cm}
\caption{Comparisons of sketch-guided image outpainting. All methods except ours suffer from the lack of the texture details and blurry boundaries of different semantic regions. The dashed boxes indicate the blurry regions.}
\label{fig1}
\end{center}
\end{figure}

\begin{figure*}[t]
\setlength{\abovecaptionskip}{0pt} 
\setlength{\belowcaptionskip}{0pt} 
\begin{center}
\subfigure[Inputs]{
    \begin{minipage}[t]{0.18\linewidth}
        \centering
        \includegraphics[width=1.34in]{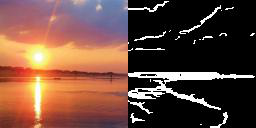}\\
        \vspace{0.05cm}
        \includegraphics[width=1.34in]{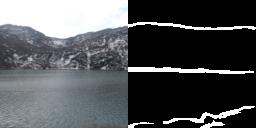}\\
        \label{ms:a}
        \vspace{0.05cm}
    \end{minipage}
    }
\subfigure[Reference Images]{
    \begin{minipage}[t]{0.18\linewidth}
        \centering
        \includegraphics[width=1.34in]{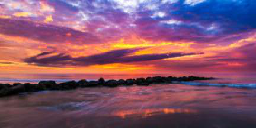}\\
        \vspace{0.05cm}
        \includegraphics[width=1.34in]{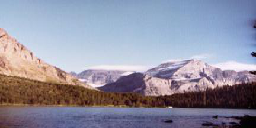}\\
        \label{ms:b}
        \vspace{0.05cm}
    \end{minipage}
    }
\subfigure[w/o references]{
    \begin{minipage}[t]{0.18\linewidth}
        \centering
        \includegraphics[width=1.34in]{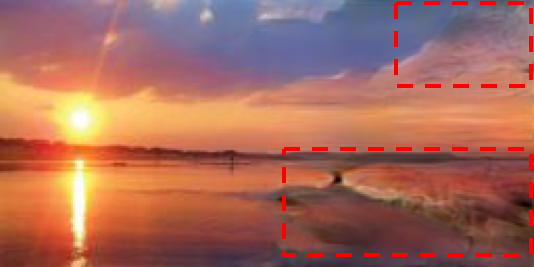}\\
        \vspace{0.05cm}
        \includegraphics[width=1.34in]{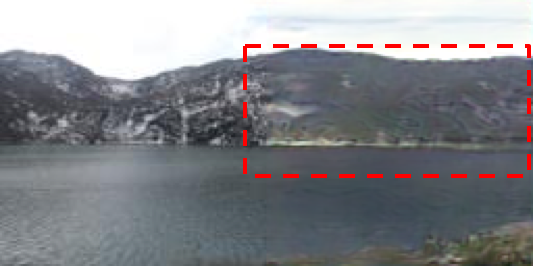}\\
        \label{ms:c}
        \vspace{0.05cm}
    \end{minipage}
    }
\subfigure[w/ references]{
    \begin{minipage}[t]{0.18\linewidth}
        \centering
        \includegraphics[width=1.34in]{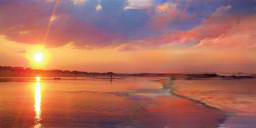}\\
        \vspace{0.05cm}
        \includegraphics[width=1.34in]{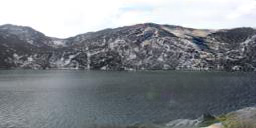}\\
        \label{ms:d}
        \vspace{0.05cm}
    \end{minipage}
    }
\subfigure[Groundtruth]{
    \begin{minipage}[t]{0.18\linewidth}
        \centering
         \includegraphics[width=1.34in]{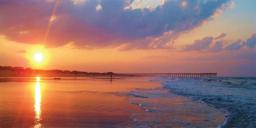}\\
         \vspace{0.05cm}
        \includegraphics[width=1.34in]{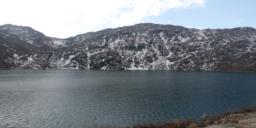}\\
        \label{ms:e}
        \vspace{0.05cm}
    \end{minipage}
    }
\caption{The outpainting examples with (w/) and without (w/o) the reference images. Although the model abandoning the reference images could predict reasonable pixels for the inputs, but its outpainting results suffer from the lack of textural details. Since the neighbor images share many pixels with the image to be extended, by borrowing some valuable pixels from the reference images, the model could synthesize outpainting results with rich texture.}
\label{ms}
\end{center}
\end{figure*}

Although existing methods could produce coherent content for the given image patch, results are still not satisfactory due to the lack of texture details. In Fig.~\ref{fig1:b}-~\ref{fig1:d}, we visualize the outpaining results produced by current state-of-the-art methods~\cite{yangzx,bds,wang}. Generally, these methods could successfully synthesize the desired images matching the guiding sketches. However, if we look closely at the details, 
many poor generated regions such as pixels with fewer texture particulars and blurry boundaries between different semantic regions can be observed. As a consequence, the overall outpainting results are not authentic enough.


Intuitively, an landscape photo is usually with similar layout and appearance to those photos in the same scene. As shown in the top row of Fig.~\ref{ms}, both the input patch and the reference image show the sunset related scene, and there are many valuable pixels in the reference image to help synthesize the high-quality content for the input patch. Therefore, if we can successfully transfer the knowledge from similar photos to complement the textural details of the predicted content, the authenticity of the generated part may be significantly improved.
Straightforwardly, the input image itself is a natural choice for serving as the reference, since it often contains content-consistent pixels to the outpainting part. However, simply adopting the input image for referring often limits the diversity of sketch layout or content pattern of the outpainting part, leading to poor generalization ability, especially for the free-form outpainting.

Motivated by the above observations and considerations, in this work, we investigate a principle to synthesize detailed outpainting results by taking the pixels from the neighbors (\ie, reference images) of the given input as the guidance, named Reference-Guided Outpainting (ReGO). Since the reference images inevitably include some inconsistent content, simply performing the transferring process without adaptive filtering will introduce abrupt pixels, 
reducing the quality of the generated part accordingly. Thus, the main challenge of ReGO lies in how to appropriately transfer pixels from neighbors while maintain the style consistent with the input image.



To this end, an Adaptive Content Selection (ACS) module is first proposed to augment our ReGO. Concretely, an image-guided convolution is first conducted on the reference image to select the compensatory features, and two feature fusion blocks are followed for guiding sketch fusing and boundary stitching, respectively. With the ACS module, our ReGO can effectively filter out the abrupt or profitless contents and only adopt the useful features to synthesize texture-rich results. Besides, the introduced reference image is only responsible for contributing contents to enrich the final outpainting results, while the style of the synthesized part should keep consistent with the input, instead of being affected by the reference. To achieve the style consistency, we further utilize a hinge-based ranking loss to pull the style of generated part close to the input image and prevent the style of generated image from biasing to the reference image. Particularly, the generated part and the input patch are treated as the positive pair, while the reference image is regarded as the negative sample, then, the triplet loss is employed to constrain their style representations. 
The style ranking loss could make our system avoid some abrupt pixels and synthesize the results with smooth style. We perform experiments on two popular benchmarks, NS6K~\cite{yangzx} and NS8K~\cite{wang}. Extensive quantitative and qualitative comparisons can well demonstrate the superiority of our ReGO over other state-of-the-art approaches.

In sum, we highlight the contributions of this work as follows:
\begin{itemize}
    \item We propose an effective outpainting method named ReGO, which can transfer content-consistent pixels from reference images to synthesize texture-rich results.
    \item A novel adaptive content selection module is designed to pick up the beneficial features from the reference image, making ReGO could filter out some abrupt pixels and generate semantic-consistent results.
    \item A style ranking loss is proposed to restrain the style of the synthesized part and enables the system to synthesize style-consistent results.
    \item State-of-the-art performance on both random outpainitng and sketch-guided outpainting tasks.
\end{itemize}

\begin{figure*}
    \centering
    \includegraphics[width=7.2in, height=2.3in]{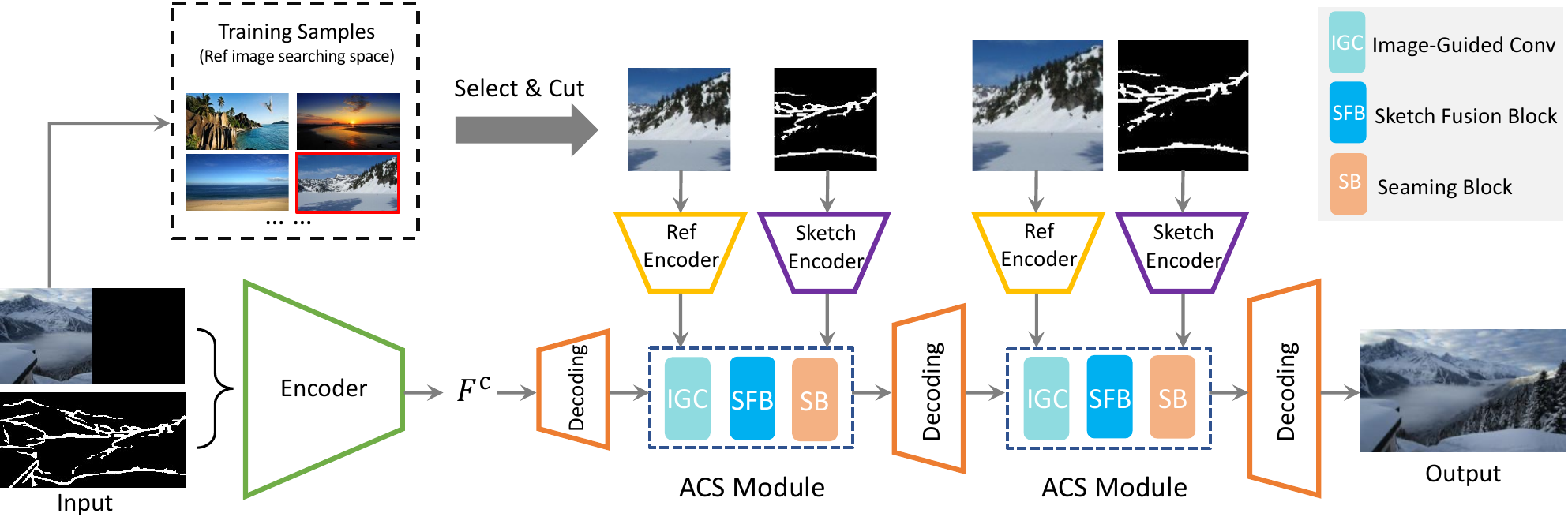}
    \caption{The overview of the proposed ReGO. The ReGO system takes the left image \& the sketch as inputs, and synthesizes the additional right half new content for the input image. The overall architecture follows an encoder-decoder paradigm, where the encoder compresses the inputs and obtain the hidden feature $F$, and the decoder is responsible to rebuild the complete image from $F$. Meanwhile, our Adaptive Content Selection (ACS) module could be equipped in each decoder layer, a reference image is first selected from the training samples, whose right half is further cut and fed into the proposed ACS module together with the guiding sketch to replenish the hidden features. As for the guiding sketch, we use the groundtruth from the right half image during training. At the testing stage, the guiding sketch could be manually drawn or borrowed from other image, as shown in Fig.\ref{free_form comparison_transfer_guide}.}
    \label{framework}
    \vspace{0.2cm}
\end{figure*}

\section{Related Work}
\noindent\textbf{Image Inpainting.}
The image inpainting has been well explored recent years, whose target is to restore the missing or corrupt regions in images~\cite{guozongyu0,Xie0,Yu0,transfill, Pathak_2016_CVPR,tip1,tip2,tip3,ict,texture_inpaint}. Benefit from the tremendous success of the deep learning and the generation adversarial networks (GANs)~\cite{GAN}, the image inpainting has made great advances these years. 
In the early exploratory stage of this task, the researchers target on the missing regions with formal shapes~\cite{GL,Yu_2018_CVPR}, the core idea is to collect information from the surrounding context to restore the missing pixels. For example, Liu \etal develop a novel operation named partial convolution, which could iteratively predict the missing pixels by collecting information from the surrounding content~\cite{Liu0}. With the technique developing, the community pays more attention to the free-form outpainting problems~\cite{Yu0,Yu_2018_CVPR,inpaint_xie,guozongyu0,Liu0,nolocal_inpaint}.  In~\cite{inpaint_xie}, Xie \etal propose a feature re-normalization to adapt to the irregular holes. In~\cite{guozongyu0}, Guo \etal propose a full-resolution residual network (FRRN) to restore the missing pixels with irregular shape. Comparing to the inpainting,  the missing pixels of outpainting task are far from the valid content, posing more challenges to restore.

\noindent\textbf{Image Outpainting.}
Conventional image outpainting methods follow a search-and-compose pipeline, where the potential patches are first selected from an external library and then stitched with the input image to conduct extrapolation~\cite{Shan_ECCV2014,mwang,zhang_cvpr2013,Efros,T_inpaint2}. For example, Wang \etal~\cite{mwang} first retrieve candidate images using subgraph matching, and stitch the wrapped images into the input by smoothing the boundary. Inspired by the success of the deep neural network, researchers recently attempt to predict new content beyond the boundaries using the generative adversarial networks (GANs)~\cite{GAN,bds,yangzx}. For example, Yang \etal~\cite{yangzx} utilize the recurrent neural network ~\cite{LSTM} to iteratively produce new content, the proposed algorithm could synthesize very long outpainting results. Teterwak \etal~\cite{bds} develop an encoder-decoder based generator to predict the unseen pixels. The corrupt image is first compressed by the encoder and restored by the decoder. However, these methods could only predict random contents, to address this weakness, the conditional image outpainting starts to be studied recently. Wang \etal~\cite{wang} develop a network that allows users to guide the final synthesis by free-form sketches. In~\cite{condition_progress}, the authors propose to utilize the language and the position clues to control the outpainting results. Nevertheless, existing methods still suffer from the blur texture of the resutls, which motivates us to develop a controllable outpainting framework that can synthesize images with rich texture.

\begin{figure*}
    \centering
    \includegraphics[width=6in, height=3in]{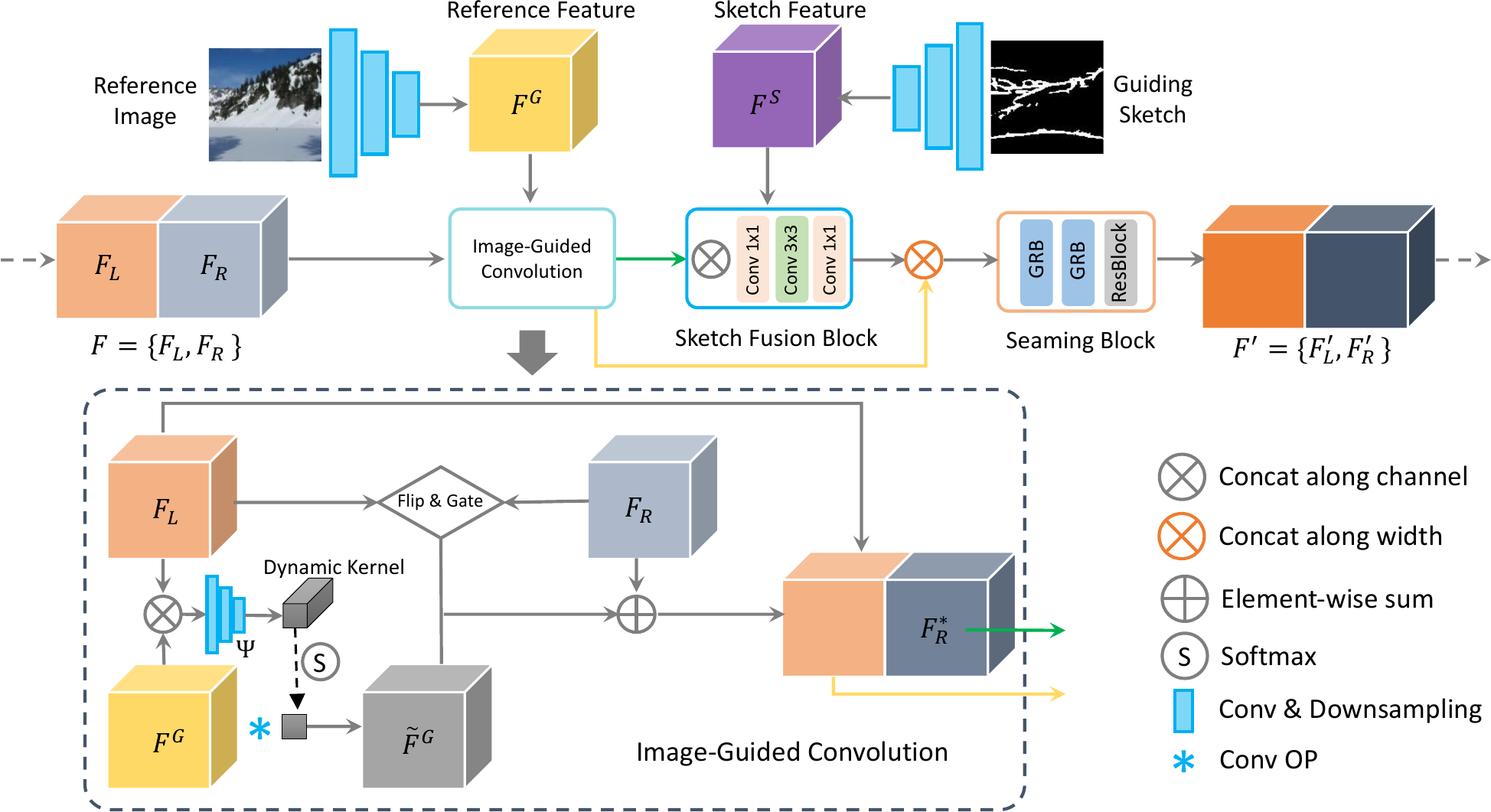}
    \caption{The architecture of our proposed ACS module. The profitable features are first distilled from the reference image, and then used to replenish the predicted new features. The sketch fusing block is followed to fuse the sketch clues, to build a controllable outpainting system. Finally, a seaming block is applied to achieve the smooth transition from the left to the right content.}
    \label{workflow}
    \vspace{-0.4cm}
\end{figure*}

\section{Methodology}
\label{method}
\noindent \textbf{The Overview.}
Fig.~\ref{framework} exhibits the pipeline of our outpainting system, whose architecture is designed in an encoder-decoder paradigm, the proposed Adaptive Content Selection (ACS) module is plugged into each decoder layer.  Particularly, the encoder takes the left half image and the sketch as inputs and outputs the hidden feature map $F$, which is fed into subsequent decoder layers to rebuild the complete image. To synthesize the texture-rich results, the reference image is first chosen from the searching space. \ie, training samples and fed into the ACS module to distill its content for compensating purpose. Besides, to allow the user to harvest the freestyle outpainting, the guiding sketch clue is also integrated to build a flexible system.
Fig.~\ref{workflow} shows the details of the ACS module. The image-guided convolution is first employed to distill the beneficial features from the reference image, then the selected features are integrated with the hidden representations of the synthesized part to complement the texture details. In addition, the style ranking loss is designed to encourage the generator to produce style-consistent content. 

Hereinafter, we first introduce the data structure used in this work in subsection~\ref{data_pre}, and then the details of proposed ACS module and style ranking loss are subsequently presented in subsection~\ref{IGConv} and subsection~\ref{SRL}, respectively.


\subsection{Data Preparation}
\label{data_pre}
For an image $I$ from training set $X$, the corresponding sketch and the reference image are two necessary auxiliary data for our system. The sketch is treated as a conditional clue to guide the synthesis procedure as shown in Fig.~\ref{fig1}, while the reference image is responsible to provide rich detailed features for the predicted new content. 

To obtain the sketch, we first extract the edge map using the HED edge detector~\cite{HED}, and binarize the generated edge map with a pre-defined threshold (0.6 as adopted in our experiment), to acquire the binary sketch $S\in \mathbb{R}^{H\times W\times 1}$. In the training stage, we adopt the original sketches of the missing parts from training samples as inputs and force the network to restore the ground-truth parts. At the testing stage, users could input the manually drawn free-form sketches to synthesize desired results. 

To obtain reference images for the input $I$, we first extract feature representations using a pre-trained Places365 \cite{places365} model, and then apply the cosine similarity to identify the similar samples. Based on our observation, the visual neighbors often share similar content with the target image and have more beneficial pixels, thus they are reasonable to be treated as the reference images. In our practice, we select multiple neighbors for each sample and randomly pick up one in each training iteration as the reference image $G$. Besides, since we only need to replenish the details of synthesized part, the right half of the reference image is only considered for further processing, as shown in Fig.~\ref{workflow}.


\subsection{ACS Module}
\label{IGConv}
With the reference image and the guiding sketch, the responsibilities of our ACS module are two-fold, \ie,  compensating the texture details of the synthesized new content and fusing the conditional information from the guiding sketch. Consequently, an image-guided convolution is designed for detail enriching, and a sketch fusing block is employed to integrate the sketch clue. Additionally, we further apply a seaming block to smooth the boundary between the original left feature and the predicted new representation. The details of each block are subsequently presented in the following.

\vspace{0.15cm}
\noindent\textbf{Image-Guided Convolution.}
The proposed Image-Guided Convolution (IGConv) aims to help the network complement texture details for the new content using the distilled the beneficial features from the reference image. \yx{Intuitively, the left part of the input could directly serve as the reference image. However, based on our experience, such a strategy often results in an insufficiently diverse training set. As a consequence, the trained model would have a deep impression on the original sketch layout and content pattern and fail to generalize to the freestyle outpainting. Therefore, in our Image-Guided Convolution, we first search multiple neighbors for the reference image and randomly select a neighbor in each iteration as the reference. Such a training fashion allows the model to see diverse input-reference pairs, thus, the generality of the final model could be boosted accordingly. }   

Formally, let $F_L \in \mathbb{R}^{h\times w\times c}$ be the features encoded from the image to be extended, $F_R$ represents the hidden features for the predicted new content. $F_L$ and $F_R$ form the complete hidden features of the overall image $F\in \mathbb{R}^{h\times 2w\times c}$. And the features of $G$, which are encoded from a reference image encoder, are denoted as $F^G\in \mathbb{R}^{h\times w\times c}$. The designed image-guided convolution aims to complement $F_R$ by extracting helpful information from reference features $F^G$. A group of dynamic filters are conditionally produced based on the features of input and reference images, making the network adaptively collect the beneficial content from the reference image.
To be specific, a dynamic kernel is produced based on the concatenation of $F^G$ and $F_L$ via a simple feed-forward procedure:
\begin{equation}
\label{dynamic_kernel}
k = \Psi(F^G, F_L),
\end{equation}
where \yx{$k\in \mathbb{R}^{3\times 3\times c}$, 3} and $c$ indicates the kernel size and channel number, respectively. \yx{The $\Psi$ can be modeled as the neural network, which takes the features of the reference and the input and recurrently use the convolution with stride=2, batch normalization and ReLU\footnote{ReLU is not applied in the last layer} to get the features.} 

The dynamic kernel in Eq.~\ref{dynamic_kernel} targets on providing guidance to distill the content of the reference image. To adaptively pick up the beneficial pixels and restrain the unhelpful content, we conduct the channel-wise normalization to update the dynamic kernel:
\begin{equation}
\label{softmax}
    k^n_{ijk} = \frac{\exp{(k_{ijk})}}{\sum_{h}{\exp {(k_{ijh)}}}}.
\end{equation}
Thus, the distilled $i$-th channel map can be obtained as follow:
\begin{equation}
\label{distill}
    \widetilde{F}^G_{:,:,i} = F^G * P(k^n_{:,:,i}), i=1,2,...,c,
\end{equation}
where the $*$ denotes the convolution operation, $P(\cdot)$ is an operation to repeat $k^n_{:,:,i}$ c times along the channel dimension. All of the $ \widetilde{F}^G_{:,:,i}|^{c}_{i=1}$ are channel-wise stacked to get the distilled feature map $\widetilde{F}^G \in \mathcal{R}^{h\times w\times c}$. 

The dynamic kernel in Eq.~\ref{softmax} focuses on automatically highlighting the beneficial features and suppressing the profitless content, we expect the network could assign lower enough weights for the profitless featurs while give higher weights for the helpful contents via the softmax operation. And the following distilled convolution in Eq.~\ref{distill} attempts to greedily summarize the profitable semantic regions across feature channels from each dynamic kernel's perspective. With such a procedure, the helpful features from the reference image are effectively aggregated into the map $\widetilde{F}^G$, which are further added to the feature $F_R$ to achieve the compensatory purpose: $F_R^{*} = F_R + \widetilde{F}^G$.  

Besides the reference image, the input image itself could also provide valuable pixels. It is because the synthesized part contains the same objects as the input image with high probability. Therefore, the features from the input image are also integrated and the $F_R^{*}$ is updated as follow:
\begin{equation}
    F_R^* = F_R + \widetilde{F}^G + F_L\times\sigma(\rho(F_L) \times F_R),
\end{equation}
where $\rho(\cdot)$ is the horizontally flip operation, $\sigma(\cdot)$ denotes the sigmoid function, which is introduced to learn a dynamic feature selection mechanism. 

\vspace{0.15cm}
\noindent\textbf{Sketch Fusing Block.}
Besides synthesizing the unseen part with thriving and realistic details, our ReGO should also be equipped with a practical mechanism, \ie, allowing users to acquire personal custom outpainting results using their preferred sketches as the guidance. To this end, we introduce a controllable sketch fusion block to achieve the target. To make the final results exactly match the guiding sketch, the sketch fusion block additionally integrates the sketch feature to emphasize the desired shape in the restoring procedure, as shown in Fig.~\ref{workflow}.

Concretely, only the right half sketch $S^r\in \mathbb{R}^{H\times W/2\times 3}$ serves as the guiding clues, and its feature maps $F^s$ are first encoded by a sketch encoder $E^S$. Then, the compressed sketch features are channel-wise concatenated with the complemented feature $F_R^{*}$, and fed forward a residual block~\cite{resBlock} style structure to get the fused output $F_R^s$. 

\vspace{0.15cm}
\noindent\textbf{Seaming Block.}
Our \yx{seaming block} is responsible to fuse the raw left half features $F_L$ and the complemented features $F_R^{s}$, which in fact attempts to smooth the boundary between the raw features from the input image and the complemented right half features. As shown in Fig.~\ref{workflow}, the seaming block consists of two global residual blocks (GRB)~\cite{yangzx} and a residual block~\cite{resBlock}. We alternately utilize the 1$\times$3 and 7$\times$1 convolution in GRB to strengthen the connection between the original and the predicted regions, especially the boundary between the map from the input image and the complemented map of the predicted new content. Particularly, the $F_L$ and $F_R^{s}$ are first concatenated along the width dimension, and then sequentially fed through two GRBs and a residual block to get the output $F^{'}$, which is also the final output of our ACS module.

\setlength{\tabcolsep}{3mm}{
\begin{table*}[t]
\centering
\caption{Performance comparisons on three datasets under criteria IS, FID and MSD, for sketch-guided and random outpainting tasks. $*$ means the method is modified to perform sketch-guided outpainting as descripted in subsection~\ref{imple}.}
\begin{tabular}{|c|ccc|ccc|ccc|}
\hline
\multirow{3}{*}{} &\multicolumn{9}{c|}{Sketch-Guided Outpainting} \\
\cline{2-10}
& \multicolumn{3}{c|}{NS6K}& \multicolumn{3}{c|}{NS8K}& \multicolumn{3}{c|}{SUN Attribute} \\
\cline{2-10}
&IS${\color{red}\uparrow}$ &FID${\color{red}\downarrow}$ &MSD${\color{red}\uparrow}$ &IS${\color{red}\uparrow}$ &FID${\color{red}\downarrow}$ &MSD${\color{red}\uparrow}$ &IS${\color{red}\uparrow}$ &FID${\color{red}\downarrow}$ & MSD${\color{red}\uparrow}$\\ 
\hline
$\text{DeepFillv2}^*$~\cite{Yu0} & 2.316 &16.712   & 0.511  & 3.012 & 15.132 & 0.592 & 9.132 & 27.993 & 0.562   \\
$\text{CoModGAN}^*$~\cite{comodgan} & 2.758 & 15.145 & 0.583  & 3.221 & 13.774 & 0.685  &  9.884 & 26.076  & 0.641   \\
$\text{LaMa}^*$~\cite{lama} &3.016 & 13.639  &  0.567 & 3.347 &12.312  & 0.629  & 10.147 & 24.135  & 0.619  \\
$\text{NSIO}^*$~\cite{yangzx} &2.891 &12.87 & 0.649 &3.254 &10.813 & 0.837 & 10.391 & 23.021 & 0.801 \\
SGIO~\cite{wang} &2.920 & 10.998 &1.01 &3.321 &10.390 & 1.17 & 10.126 & 22.536 &  0.792 \\
$\text{BDIE}^*$~\cite{bds} &3.002 & 11.021 &0.963 &3.323 &9.639 & 0.892 & 10.857 & 21.412 & 0.886 \\
\hline
$\text{ReGO}_\text{NSIO}$& 2.923 & 12.03 & 0.839 & 3.329 & 10.232 & 0.966 & 10.683 & 21.452 & 0.869\\
$\text{ReGO}_\text{SGIO}$ & 2.924 & 10.104 & 1.201 &3.387 & 9.787& 1.293  & 10.542 & 21.012 & 0.894 \\
$\text{ReGO}_\text{BDIE}$ & \bf{3.126} & \bf{10.052} & \bf{1.357} &\bf{3.444} &\bf{8.738} & \bf{1.396} &\bf{10.992} & \bf{19.433} & \bf{1.014}\\
\hline
\multirow{3}{*}{} &\multicolumn{9}{c|}{Random Outpainting} \\
\cline{1-10}
$\text{DeepFillv2}$~\cite{Yu0} & 2.273 & 18.693   &  - & 2.816 & 18.109  & - & 8.987  & 29.016  & -  \\
$\text{CoModGAN}$~\cite{comodgan} & 2.612 & 18.014 & -  & 2.983 & 17.223  & -  & 9.763  & 27.973  & -  \\
$\text{LaMa}$~\cite{lama} &2.773 & 15.061  & -  & 3.019 & 15.014 &-  &  9.974 &  25.863 & - \\
NSIO~\cite{yangzx} &2.883 &13.612 & - &3.123 &12.871 & - & 10.229 & 25.271 &- \\
SGIO~\cite{wang} &2.951 & 15.857 & - &3.178 &12.316 & - & 9.889 & 24.978 & -  \\
BDIE~\cite{bds} &2.880 & 13.252 & - &3.155 & 11.373 & - & 10.647 & 24.465 & - \\
\hline
$\text{ReGO}_\text{NSIO}$& 2.792 & 14.224 & - & 2.948 & 13.230 & - & 10.553 & 26.196 & -\\
$\text{ReGO}_\text{SGIO}$ & 2.848 & 15.396 & - &3.204 & 13.748 & -  & 10.438 & 25.441 & - \\
$\text{ReGO}_\text{BDIE}$ & \bf{3.243} & \bf{12.606} & - &\bf{3.573} &\bf{10.586} & - &\bf{10.796} & \bf{23.209} & -\\
\hline
\end{tabular}
\label{main_perform}
\end{table*}
}

\subsection{Style Ranking Loss}
\label{SRL}
The reference image adopted by ReGO is only expected to contribute texture details, its style should not be reflected on the synthesized content. 
\yx{To reduce the artifacts, the model should 1) only transfer the texture details from the reference image, 2) keep a consistent style between the given part and the generated part. }
Given the above considerations, we find the hinge based ranking is quite in line with our requirement, we take the synthesized part and the input image as the positive pair while treat the reference image as the negative sample, then the hinge ranking loss could conduct on their style representations to enforce the style of the input and the new content to be closer than the reference image. 

\begin{figure*}[t]
\setlength{\abovecaptionskip}{0pt} 
\setlength{\belowcaptionskip}{0pt} 
\begin{center}
\subfigure{
    \begin{minipage}[t]{0.18\linewidth}
        \centering
        \includegraphics[width=1.4in]{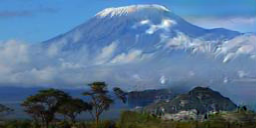}\\
        \vspace{0.05cm}
         \includegraphics[width=1.4in]{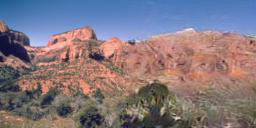}\\
        \label{rebuild:a}
        \vspace{-0.35cm}
        \mycaption{(a) BDIE~\cite{bds}}
    \end{minipage}
    }
\subfigure{
    \begin{minipage}[t]{0.18\linewidth}
        \centering
        \includegraphics[width=1.4in]{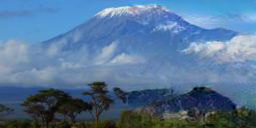}\\
        \vspace{0.05cm}
         \includegraphics[width=1.4in]{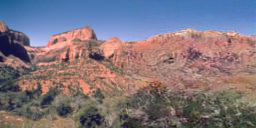}\\
        \label{rebuild:b}
        \vspace{-0.35cm}
        \mycaption{(b) +RI}
    \end{minipage}
    }
\subfigure{
    \begin{minipage}[t]{0.18\linewidth}
        \centering
        \includegraphics[width=1.4in]{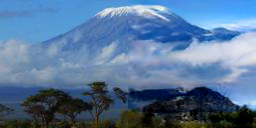}\\
        \vspace{0.05cm}
         \includegraphics[width=1.4in]{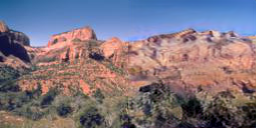}\\
        \label{rebuild:c}
        \vspace{-0.35cm}
        \mycaption{(c) +SRL}
    \end{minipage}
    }
\subfigure{
    \begin{minipage}[t]{0.18\linewidth}
        \centering
         \includegraphics[width=1.4in]{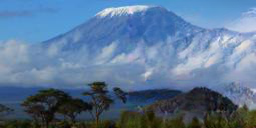}\\
         \vspace{0.05cm}
         \includegraphics[width=1.4in]{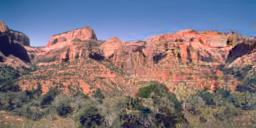}\\
        \label{rebuild:d}
        \vspace{-0.35cm}
        \mycaption{(d) +ACS}
    \end{minipage}
    }
\subfigure{
    \begin{minipage}[t]{0.18\linewidth}
        \centering
         \includegraphics[width=1.4in]{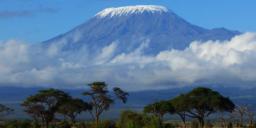}\\
         \vspace{0.05cm}
         \includegraphics[width=1.4in]{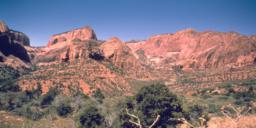}\\
        \label{rebuild:e}
        \vspace{-0.35cm}
        \mycaption{(e) Groundtruth}
    \end{minipage}
    }
    \\ \centering\text{\normalsize I: Visual ablation on image rebuilding.}\\
    \vspace{0.1cm}
\subfigure{
    \begin{minipage}[t]{0.18\linewidth}
        \centering
        \includegraphics[width=1.4in]{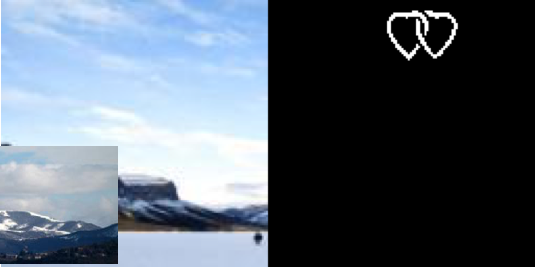}\\
        \vspace{0.05cm}
         \includegraphics[width=1.4in]{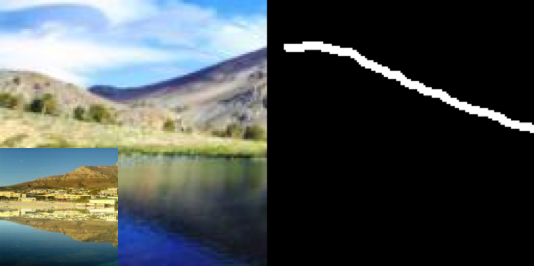}\\
        \label{free:a}
        \vspace{-0.35cm}
        \mycaption{(a) Inputs}
    \end{minipage}
    }
\subfigure{
    \begin{minipage}[t]{0.18\linewidth}
        \centering
        \includegraphics[width=1.4in]{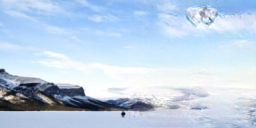}\\
        \vspace{0.05cm}
         \includegraphics[width=1.4in]{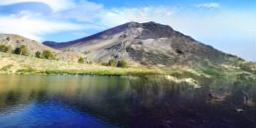}\\
        \label{free:b}
        \vspace{-0.35cm}
        \mycaption{(b) BDIE~\cite{bds}}
    \end{minipage}
    }
\subfigure{
    \begin{minipage}[t]{0.18\linewidth}
        \centering
        \includegraphics[width=1.4in]{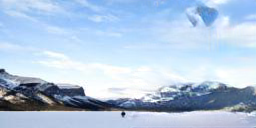}\\
        \vspace{0.05cm}
         \includegraphics[width=1.4in]{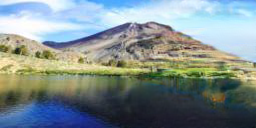}\\
        \label{free:c}
        \vspace{-0.35cm}
        \mycaption{(c) +RI}
    \end{minipage}
    }
\subfigure{
    \begin{minipage}[t]{0.18\linewidth}
        \centering
        \includegraphics[width=1.4in]{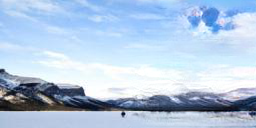}\\
        \vspace{0.05cm}
         \includegraphics[width=1.4in]{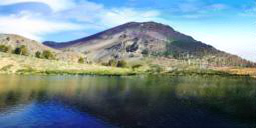}\\
        \label{free:d}
        \vspace{-0.35cm}
        \mycaption{(d) +SRL}
    \end{minipage}
    }
\subfigure{
    \begin{minipage}[t]{0.18\linewidth}
        \centering
        \includegraphics[width=1.4in]{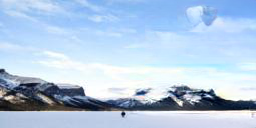}\\
        \vspace{0.05cm}
         \includegraphics[width=1.4in]{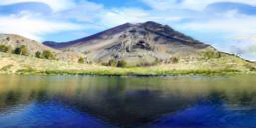}\\
        \label{free:e}
        \vspace{-0.35cm}
        \mycaption{(e) +ACS}
    \end{minipage}
    }
    \\ \centering\text{\normalsize II: Visual ablation on free-form outpainting.}\\
    \vspace{0.4cm}
\caption{The visual abaltion of each component in our method on image rebuilding and free-form outpainting, where RI, SRL, and ACS indicate the reference image, styel ranking loss, and adaptive content selection module, respectively.  }
\vspace{-0.4cm}
\label{ablation_rebuilding_and_free_form}
\end{center}
\end{figure*}

Following previous practices~\cite{style_tf1,style_tf3,style_tf2}, we utilize the second-order statistics of convolutional feature as style representation. Particularly, the style features of generated part $\hat{I}^r\in \mathbb{R}^{H\times W/2\times 3}$, which is only the right half of the image reconstruction $\hat{I}\in \mathbb{R}^{H\times W\times 3}$, is given by the Gram matrix $R^d\in \mathcal{R}^{N_{d}\times N_{d}}$:
\begin{equation}
\label{gram matrix}
    R^d_{ij} = \sum_{k} M^d_{ik}M^d_{jk},
\end{equation}
where $M^d_i$ is the vectorised $i$-th feature map in layer $d$ from a convolutional neural network like VGG19~\cite{vgg19}, \yx{$N_d$ indicates the channel number of layer $d$.}

Analogously, the style representations of the reference image and the input image can be extracted, and our style ranking loss is defined as:
\begin{equation}
    \mathcal{L}^d_s = [\alpha - SM(R^d, L^d) + SM(R^d, G^d)]_{+},
\end{equation}
where $L^d$ and $G^d$ represent the Gram matrixs of the left input and the reference image, respectively,  $SM(\cdot, \cdot)$ is the cosine similarity, $\alpha\in \mathcal{R}$ is the scalar margin, and $[\cdot]_{+} = max(\cdot, 0)$.

\begin{figure}[t]
\setlength{\abovecaptionskip}{8pt} 
\setlength{\belowcaptionskip}{8pt} 
    \centering
    \includegraphics[height=2.2in, width=3in]{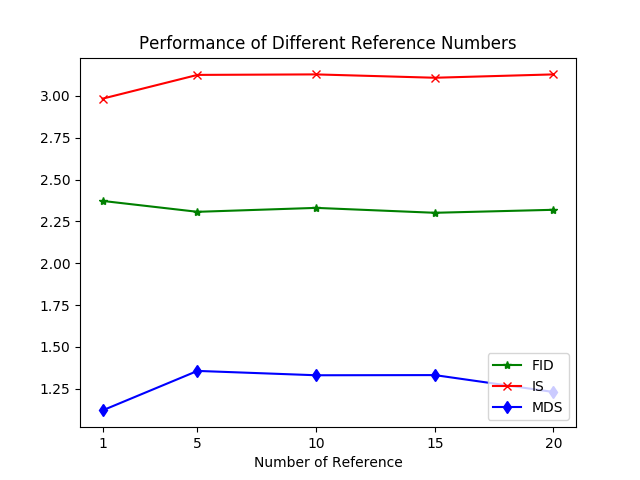}
    \caption{The IS, FID and MSD tendencies of different reference numbers, where the x-axis indicates the reference number and the y-axis is the corresponding values. We scale the FID by logarithmic function to make the value of three criteria close and exhibit clearer tendency.}
    \vspace{-0.3cm}
    \label{dis_refnum}
\end{figure}

\begin{figure*}[t]
\setlength{\abovecaptionskip}{0pt} 
\setlength{\belowcaptionskip}{0pt} 
\begin{center}
\subfigure{
    \begin{minipage}[t]{0.18\linewidth}
        \centering
        \includegraphics[width=1.4in]{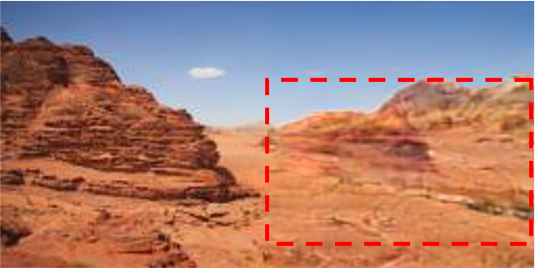}\\
        \vspace{0.05cm}
        \includegraphics[width=1.4in]{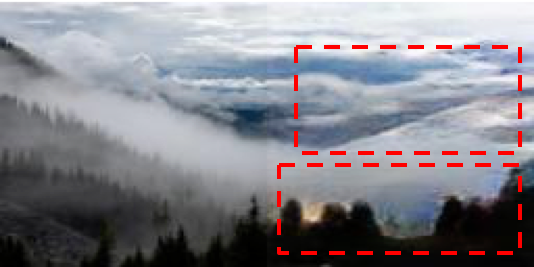}\\
        \vspace{-0.35cm}
        \mycaption{(a) LaMa~\cite{lama}}
    \end{minipage}
    }
    \subfigure{
    \begin{minipage}[t]{0.18\linewidth}
        \centering
        \includegraphics[width=1.4in]{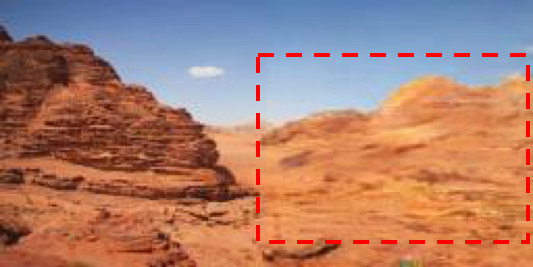}\\
        \vspace{0.05cm}
        \includegraphics[width=1.4in]{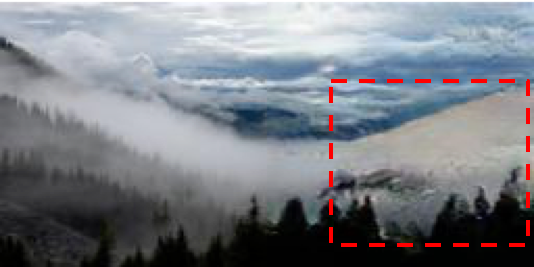}\\
        \vspace{-0.35cm}
        \mycaption{(b) SGIO~\cite{wang}}
    \end{minipage}
    }
\subfigure{
    \begin{minipage}[t]{0.18\linewidth}
        \centering
        \includegraphics[width=1.4in]{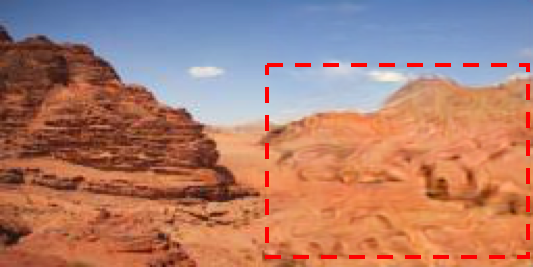}\\
          \vspace{0.05cm}
        \includegraphics[width=1.4in]{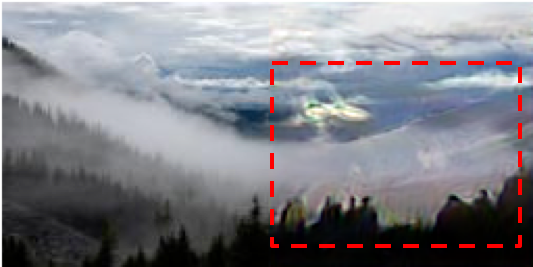}\\
        \vspace{-0.35cm}
        \mycaption{(c) BDIE~\cite{bds}}
    \end{minipage}
    }
\subfigure{
    \begin{minipage}[t]{0.18\linewidth}
        \centering
        \includegraphics[width=1.4in]{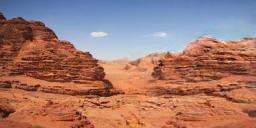}\\
        \vspace{0.05cm}
        \includegraphics[width=1.4in]{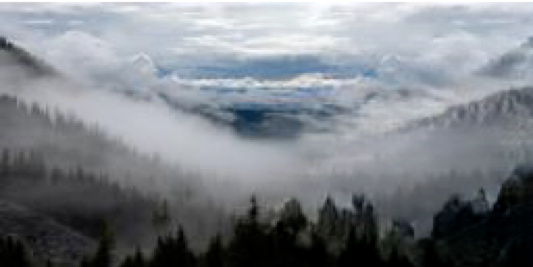}\\
        \vspace{-0.35cm}
        \mycaption{(d) $\text{ReGO}_\text{BDIE}$}
    \end{minipage}
    }
\subfigure{
    \begin{minipage}[t]{0.18\linewidth}
        \centering
        \includegraphics[width=1.4in]{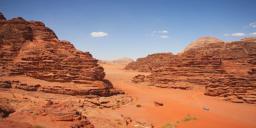}\\
        \vspace{0.05cm}
        \includegraphics[width=1.4in]{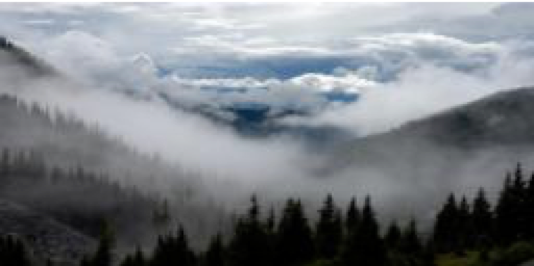}\\
        \vspace{-0.35cm}
        \mycaption{(e) Groundtruth}
    \end{minipage}
    }
    \\
    \centering\text{\normalsize I: Image rebuilding according to the original sketches.} \\
    \vspace{0.1cm}
\subfigure{
    \begin{minipage}[t]{0.18\linewidth}
        \centering
        \includegraphics[width=1.4in]{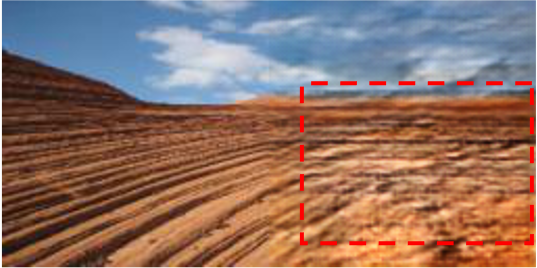}\\
        \vspace{0.05cm}
        \includegraphics[width=1.4in]{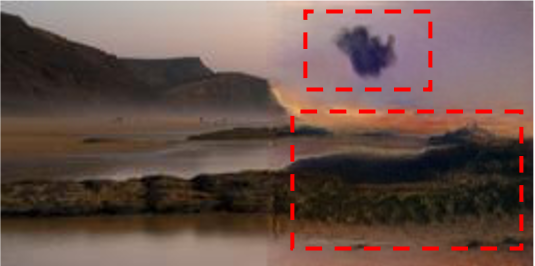}\\
        \label{random_rebuild:a}
        \vspace{-0.35cm}
        \mycaption{(a) LaMa~\cite{lama}}
    \end{minipage}
    }
    \subfigure{
    \begin{minipage}[t]{0.18\linewidth}
        \centering
        \includegraphics[width=1.4in]{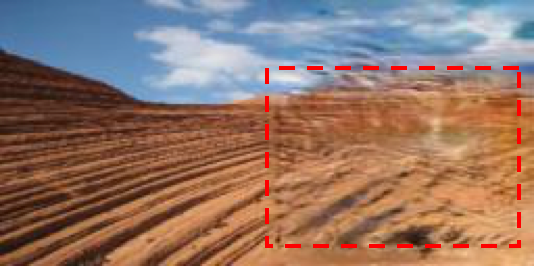}\\
        \vspace{0.05cm}
        \includegraphics[width=1.4in]{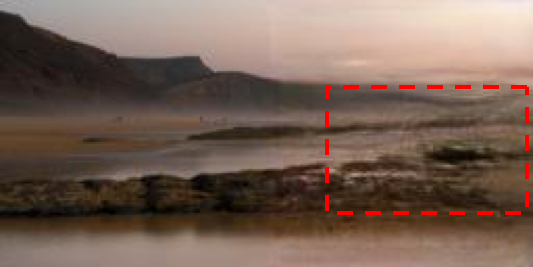}\\
        \label{random_rebuild:c}
        \vspace{-0.35cm}
        \mycaption{(b) SGIO~\cite{wang}}
    \end{minipage}
    }
\subfigure{
    \begin{minipage}[t]{0.18\linewidth}
        \centering
        \includegraphics[width=1.4in]{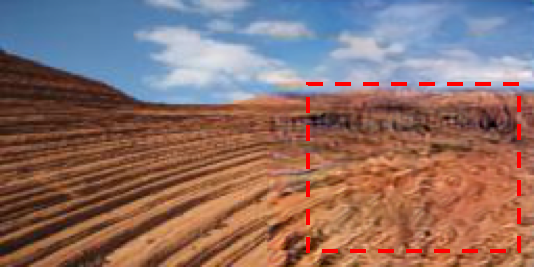}\\
        \vspace{0.05cm}
        \includegraphics[width=1.4in]{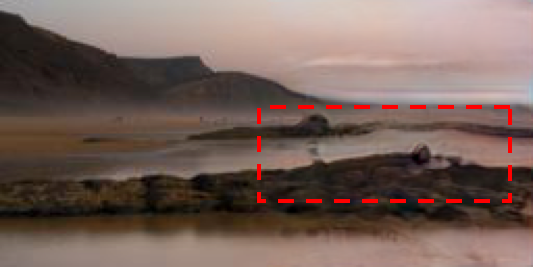}\\
        \label{random_rebuild:b}
        \vspace{-0.35cm}
        \mycaption{(c) BDIE~\cite{bds}}
    \end{minipage}
    }
\subfigure{
    \begin{minipage}[t]{0.18\linewidth}
        \centering
        \includegraphics[width=1.4in]{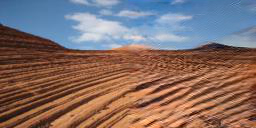}\\
        \vspace{0.05cm}
        \includegraphics[width=1.4in]{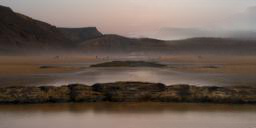}\\
        \label{random_rebuild:d}
       \vspace{-0.35cm}
        \mycaption{(d) $\text{ReGO}_\text{BDIE}$}
    \end{minipage}
    }
\subfigure{
    \begin{minipage}[t]{0.18\linewidth}
        \centering
        \includegraphics[width=1.4in]{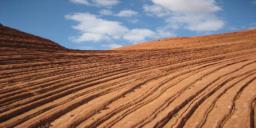}\\
        \vspace{0.05cm}
        \includegraphics[width=1.4in]{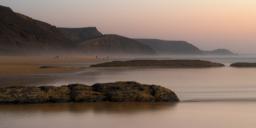}\\
        \label{random_rebuild:e}
       \vspace{-0.35cm}
        \mycaption{(e) Groundtruth}
    \end{minipage}
    }
    \\ \centering\text{\normalsize II: Random outpainting} \\
    \vspace{0.4cm}
\caption{The results for the image rebuilding and the random outpainting, where the dotted red line indicates the imperfect region. Part I shows the results on image rebuilding according to the original sketches. The comparison methods could predict reasonable pixels for the input, however, they all suffer from the blurry synthesized content. While our methods could synthesize texture-rich content. The part II exhibits the random outpainting.  Even though no sketch is provided for guidance, $\text{ReGO}_\text{BDIE}$ could also synthesise texture-rich results and performs much better than the methods originally designed for random outpainting, \ie, NSIO~\cite{yangzx} and BDIE~\cite{bds}.Red dashed frames indicate the blurry regions.}
\label{rebuiling_and_RD_outpainting}
\end{center}
\end{figure*}

\begin{figure*}[t]
\setlength{\abovecaptionskip}{0pt} 
\setlength{\belowcaptionskip}{0pt} 
\begin{center}
\subfigure{
    \begin{minipage}[t]{0.18\linewidth}
        \centering
         \includegraphics[width=1.4in]{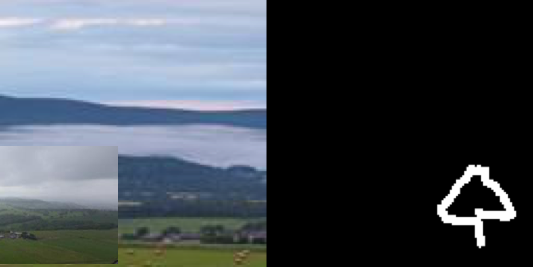}\\
         \vspace{0.05cm}
        \includegraphics[width=1.4in]{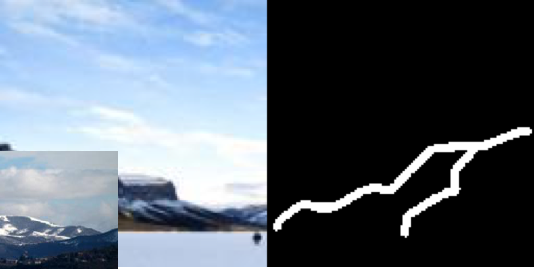}\\
        \vspace{0.05cm}
        \label{fig4:a}
        \vspace{-0.35cm}
        \mycaption{(a) Inputs}
    \end{minipage}
    }
\subfigure{
    \begin{minipage}[t]{0.18\linewidth}
        \centering
        \includegraphics[width=1.4in]{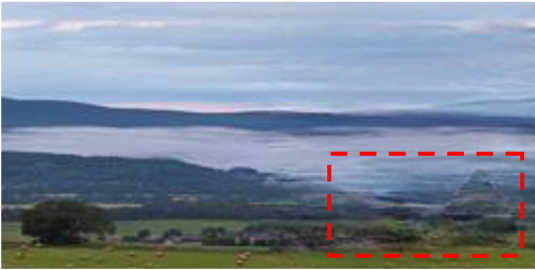}\\
        \vspace{0.05cm}
        \includegraphics[width=1.4in]{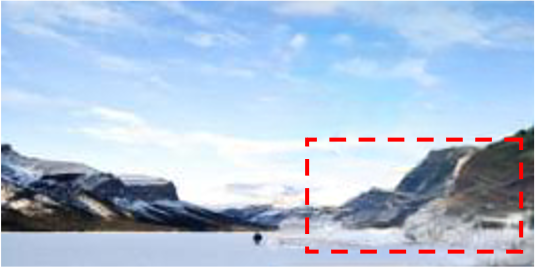}\\
        \label{fig4:b}
        \vspace{-0.35cm}
        \mycaption{(b) LaMa~\cite{lama}}
    \end{minipage}
    }
\subfigure{
    \begin{minipage}[t]{0.18\linewidth}
        \centering
        \includegraphics[width=1.4in]{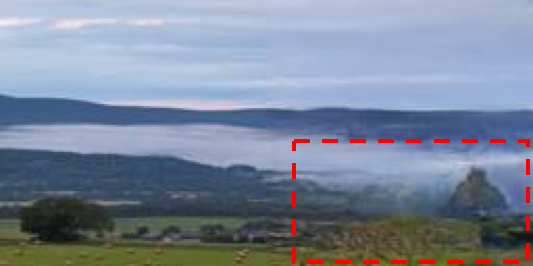}\\
        \vspace{0.05cm}
        \includegraphics[width=1.4in]{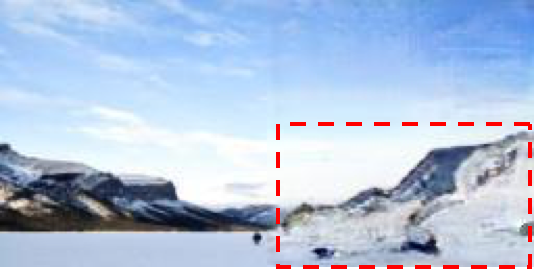}\\
        \label{fig4:c}
        \vspace{-0.35cm}
        \mycaption{(c) SGIO~\cite{wang}}
    \end{minipage}
    }
\subfigure{
    \begin{minipage}[t]{0.18\linewidth}
        \centering
         \includegraphics[width=1.4in]{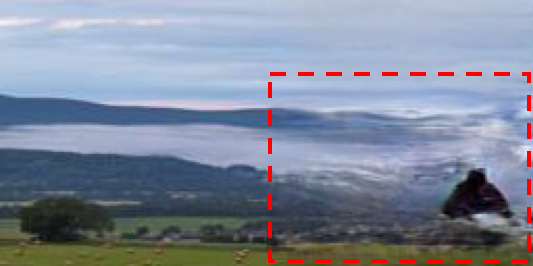}\\
         \vspace{0.05cm}
        \includegraphics[width=1.4in]{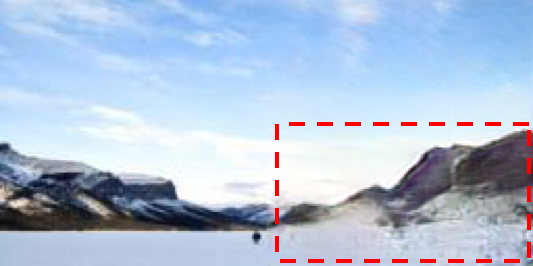}\\
        \label{fig4:d}
        \vspace{-0.35cm}
        \mycaption{(d) BDIE~\cite{bds}}
    \end{minipage}
    }
\subfigure{
    \begin{minipage}[t]{0.18\linewidth}
        \centering
        \includegraphics[width=1.4in]{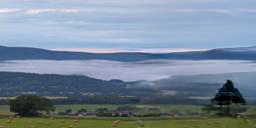}\\
        \vspace{0.05cm}
        \includegraphics[width=1.4in]{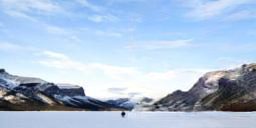}\\
        \label{fig4:e}
        \vspace{-0.35cm}
        \mycaption{(e) $\text{ReGO}_\text{BDIE}$}
    \end{minipage}
    }\\
     \centering\text{\normalsize I: Outpainting using manually drawn free-form sketches. }\\
     \vspace{0.1cm}
\subfigure{
    \begin{minipage}[t]{0.18\linewidth}
        \centering
         \includegraphics[width=1.4in]{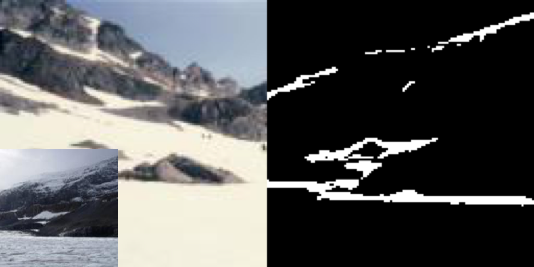}\\
         \vspace{0.05cm}
        \includegraphics[width=1.4in]{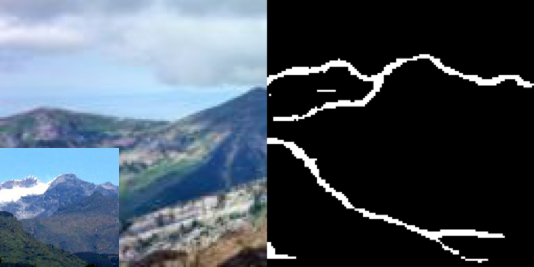}\\
        \label{detail_guide:a}
        \vspace{-0.35cm}
        \mycaption{(a) Inputs}
    \end{minipage}
    }
\subfigure{
    \begin{minipage}[t]{0.18\linewidth}
        \centering
        \includegraphics[width=1.4in]{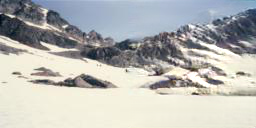}\\
        \vspace{0.05cm}
        \includegraphics[width=1.4in]{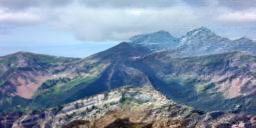}\\
        \label{detail_guide:b}
        \vspace{-0.35cm}
        \mycaption{(b) Outputs}
    \end{minipage}
    }
\subfigure{
    \begin{minipage}[t]{0.18\linewidth}
        \centering
        \includegraphics[width=1.4in]{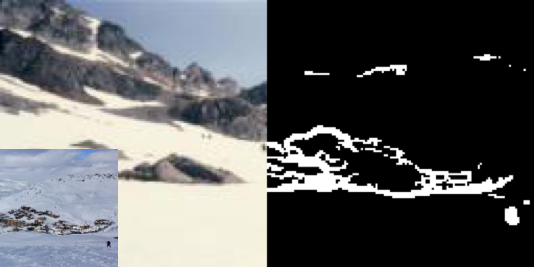}\\
        \vspace{0.05cm}
        \includegraphics[width=1.4in]{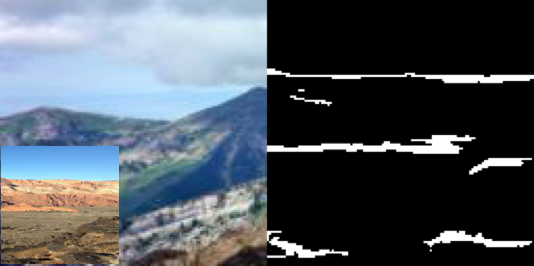}\\
        \label{detail_guide:c}
        \vspace{-0.35cm}
        \mycaption{(c) Inputs}
    \end{minipage}
    }
\subfigure{
    \begin{minipage}[t]{0.18\linewidth}
        \centering
         \includegraphics[width=1.4in]{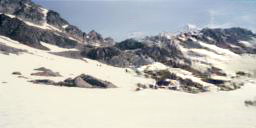}\\
         \vspace{0.05cm}
        \includegraphics[width=1.4in]{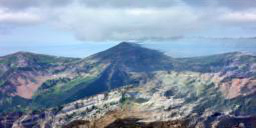}\\
        \label{detail_guide:d}
        \vspace{-0.35cm}
        \mycaption{(d) Outputs}
    \end{minipage}
    }
\subfigure{
    \begin{minipage}[t]{0.18\linewidth}
        \centering
        \includegraphics[width=1.4in]{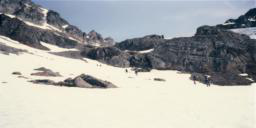}\\
        \vspace{0.05cm}
        \includegraphics[width=1.4in]{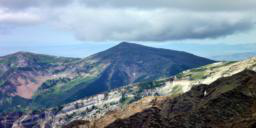}\\
        \label{detail_guide:e}
        \vspace{-0.35cm}
        \mycaption{(e) Original images}
    \end{minipage}
    }
    \\
    \text{\normalsize II: Outpainting using sketches from the other images as guidance. }
    \vspace{0.4cm}
\caption{Outpainting results according to the manually drawn free-form sketches and the sketch from other images, where the dotted red line indicates the imperfect region. Part (I) shows the results for free-form outpainting. While the part (II) exhibits the outpainting using sketches from the other images as guidance. The inputs in (a) and (c) use the same reference images (lower left in (a)) but different sketches.
The sketches in (a) are directly from the reference images, while the sketches in (c) are extracted from a randomly selected image (shown in lower left). The corresponding predictions are shown in (b) and (d), respectively. The results are produced by $\text{ReGO}_\text{BDIE}$. Red dashed frames indicate the blurry regions.}
\label{free_form comparison_transfer_guide}
\end{center}
\end{figure*}

\begin{figure*}[t]
\setlength{\abovecaptionskip}{0pt} 
\setlength{\belowcaptionskip}{0pt} 
\begin{center}
\subfigure{
    \begin{minipage}[t]{0.18\linewidth}
        \centering
        \includegraphics[width=1.34in]{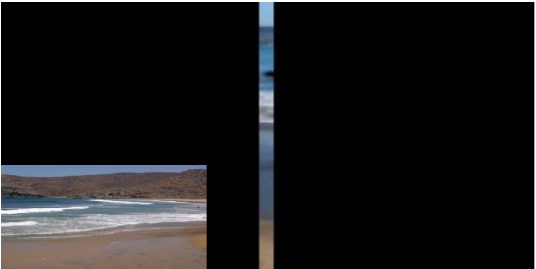}\\
        \vspace{0.05cm}
        \includegraphics[width=1.34in]{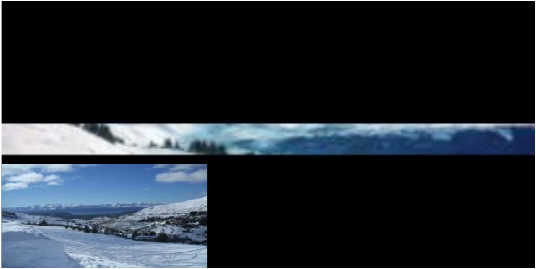}\\
        \vspace{0.05cm}
        \includegraphics[width=1.34in]{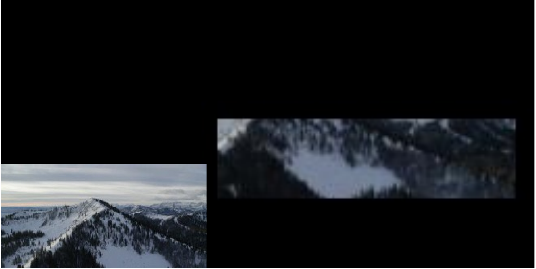}\\
        \label{fig3:a}
        \vspace{-0.35cm}
        \mycaption{(a) Input}
    \end{minipage}
    }
    \subfigure{
    \begin{minipage}[t]{0.18\linewidth}
        \centering
        \includegraphics[width=1.34in]{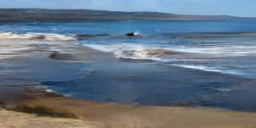}\\
        \vspace{0.05cm}
        \includegraphics[width=1.34in]{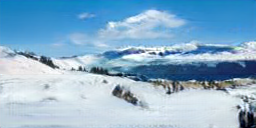}\\
        \vspace{0.05cm}
        \includegraphics[width=1.34in]{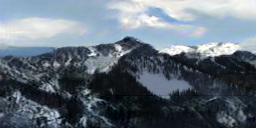}\\
        \label{fig3:c}
        \vspace{-0.35cm}
        \mycaption{(b) Output}
    \end{minipage}
    }
\subfigure{
    \begin{minipage}[t]{0.18\linewidth}
        \centering
        \includegraphics[width=1.34in]{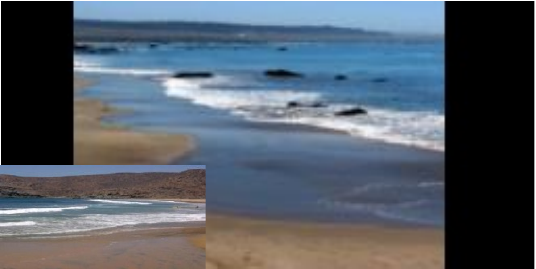}\\
        \vspace{0.05cm}
        \includegraphics[width=1.34in]{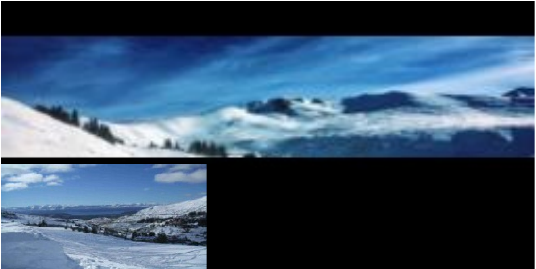}\\
        \vspace{0.05cm}
        \includegraphics[width=1.34in]{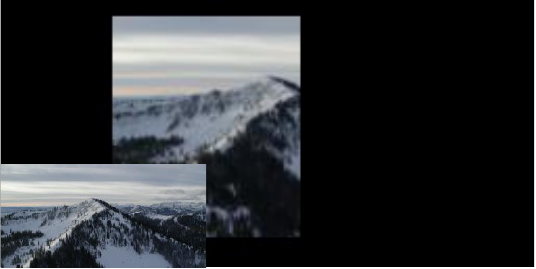}\\
        \label{fig3:b}
        \vspace{-0.35cm}
        \mycaption{(c) Input}
    \end{minipage}
    }
\subfigure{
    \begin{minipage}[t]{0.18\linewidth}
        \centering
        \includegraphics[width=1.34in]{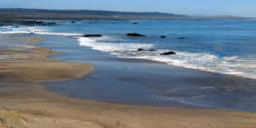}\\
        \vspace{0.05cm}
        \includegraphics[width=1.34in]{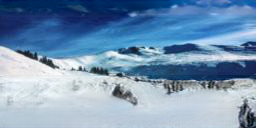}\\
        \vspace{0.05cm}
        \includegraphics[width=1.34in]{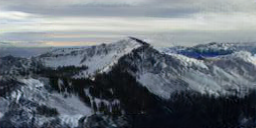}\\
        \label{fig3:d}
        \vspace{-0.35cm}
        \mycaption{(d) Output}
    \end{minipage}
    }
\subfigure{
    \begin{minipage}[t]{0.18\linewidth}
        \centering
        \includegraphics[width=1.34in]{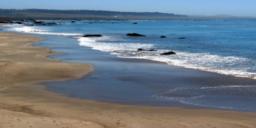}\\
        \vspace{0.05cm}
        \includegraphics[width=1.34in]{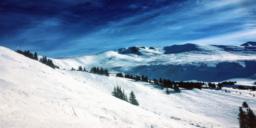}\\
        \vspace{0.05cm}
        \includegraphics[width=1.34in]{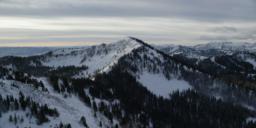}\\
        \label{fig3:e}
        \vspace{-0.35cm}
        \mycaption{(e) Groundtruth}
    \end{minipage}
    }
\caption{The results for multi-direction prediction. Based on the BDIE backbone, our method $\text{ReGO}_\text{BDIE}$ could predict the content for multiple directions.}
\vspace{-0.4cm}
\label{full_direction}
\end{center}
\end{figure*}

\begin{figure*}[t]
\begin{center}
\subfigure[LaMa~\cite{lama}]{
    \begin{minipage}[t]{0.23\linewidth}
        \centering
        \includegraphics[width=1.74in]{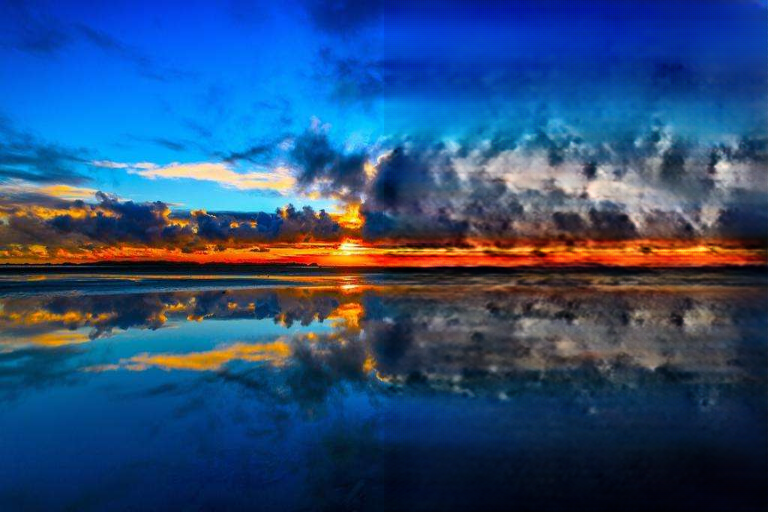}\\
        \vspace{0.1cm}
        \includegraphics[width=1.74in]{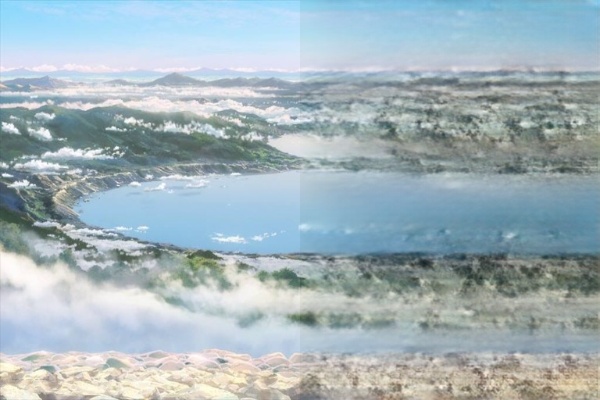}\\
        \vspace{0.2cm}
    \end{minipage}
    }
\subfigure[BDIE~\cite{bds}]{
    \begin{minipage}[t]{0.23\linewidth}
        \centering
        \includegraphics[width=1.74in]{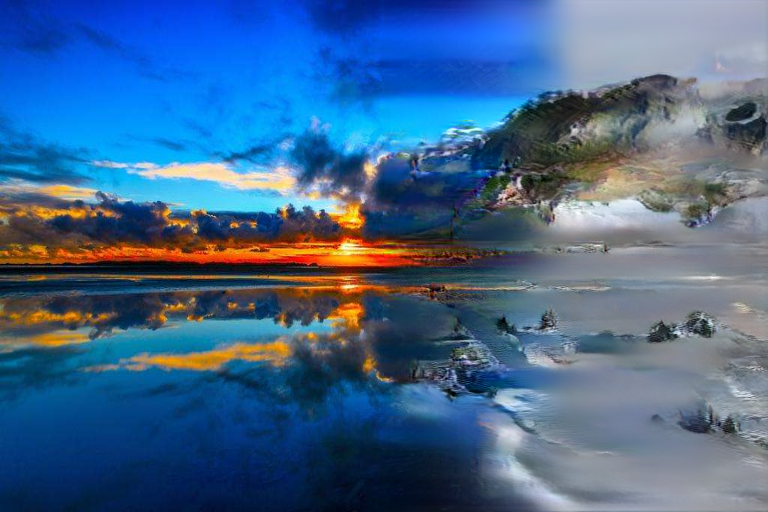}\\
        \vspace{0.1cm}
        \includegraphics[width=1.74in]{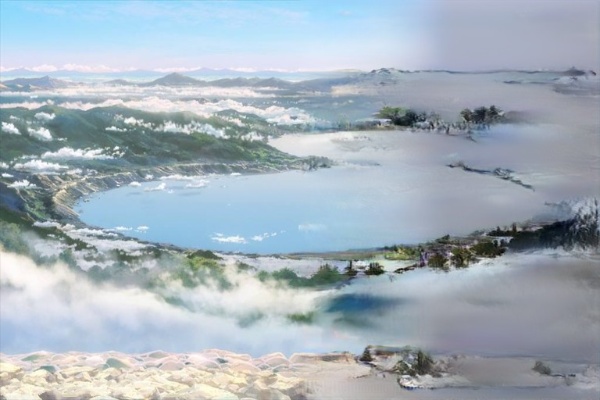}\\
        \vspace{0.2cm}
    \end{minipage}
    }
\subfigure[$\text{ReGO}_\text{BDIE}$]{
    \begin{minipage}[t]{0.23\linewidth}
        \centering
        \includegraphics[width=1.74in]{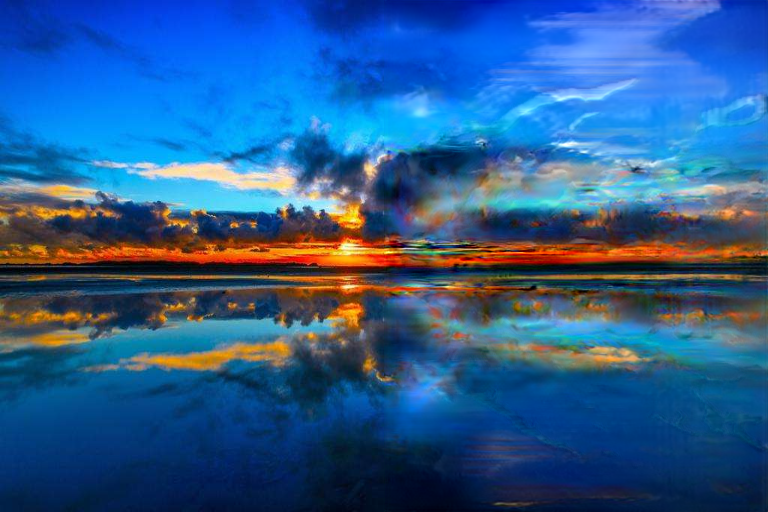}\\
        \vspace{0.1cm}
        \includegraphics[width=1.74in]{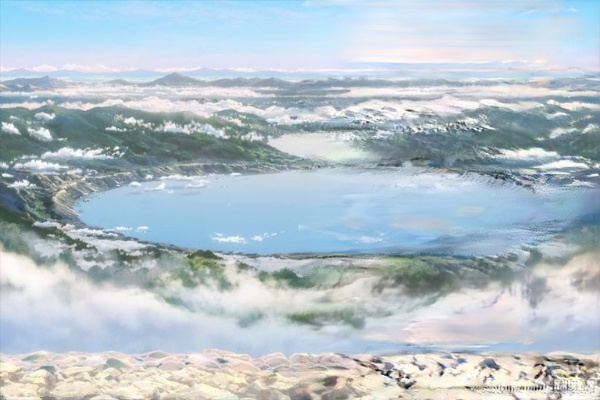}\\
        \vspace{0.2cm}
    \end{minipage}
    }
\subfigure[GroundTruth]{
    \begin{minipage}[t]{0.23\linewidth}
        \centering
        \includegraphics[width=1.74in]{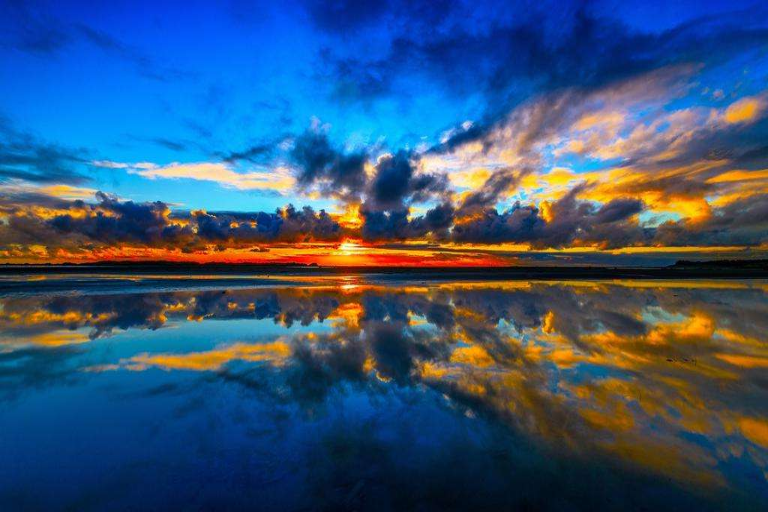}\\
        \vspace{0.1cm}
        \includegraphics[width=1.74in]{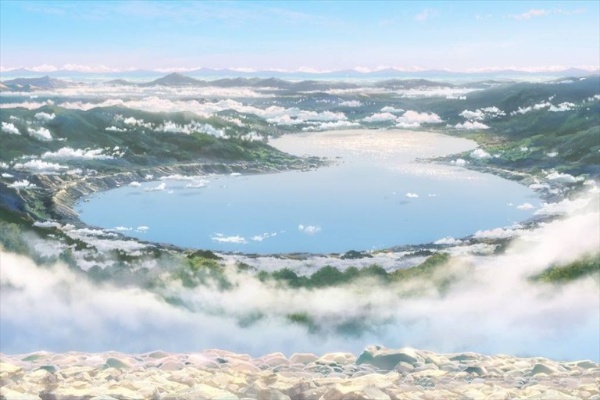}\\
        \vspace{0.2cm}
    \end{minipage}
    }
\vspace{-0.1cm}
\caption{The high resolution results of all methods, we input the images with 512$\times$384 to rebuild  512$\times$768 images.}
\label{highreso}
\vspace{-0.2cm}
\end{center}
\end{figure*}

\begin{figure*}[t]
\begin{center}
\subfigure[Inputs]{
    \begin{minipage}[t]{0.22\linewidth}
        \centering
        \includegraphics[width=1.64in]{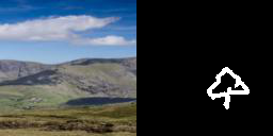}\\
        \includegraphics[width=1.64in]{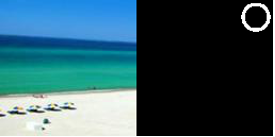}\\
        \includegraphics[width=1.64in]{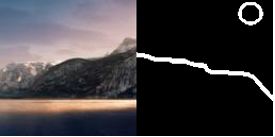}\\
         \includegraphics[width=1.64in]{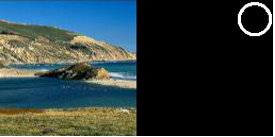}\\
        \vspace{0.1cm}
    \end{minipage}
    }
\subfigure[Outputs]{
    \begin{minipage}[t]{0.22\linewidth}
        \centering
        \includegraphics[width=1.64in]{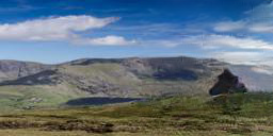}\\
        \includegraphics[width=1.64in]{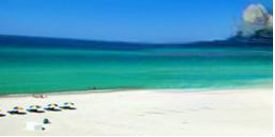}\\
         \includegraphics[width=1.64in]{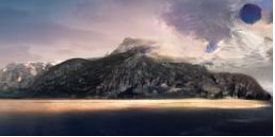}\\
        \includegraphics[width=1.64in]{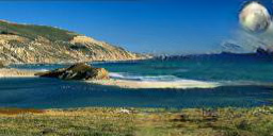}\\
          \vspace{0.1cm}
    \end{minipage}
    }
\subfigure[Inputs]{
    \begin{minipage}[t]{0.22\linewidth}
        \centering
        \includegraphics[width=1.64in]{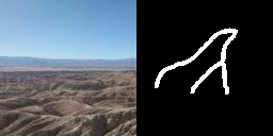}\\
        \includegraphics[width=1.64in]{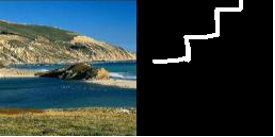}\\
         \includegraphics[width=1.64in]{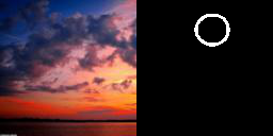}\\
        \includegraphics[width=1.64in]{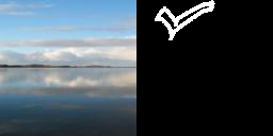}\\
          \vspace{0.1cm}
    \end{minipage}
    }
\subfigure[Outputs]{
    \begin{minipage}[t]{0.22\linewidth}
        \centering
        \includegraphics[width=1.64in]{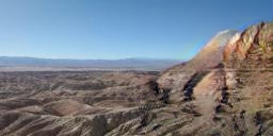}\\
        \includegraphics[width=1.64in]{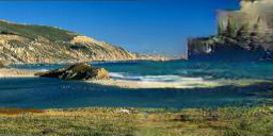}\\
         \includegraphics[width=1.64in]{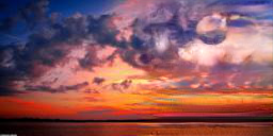}\\
        \includegraphics[width=1.64in]{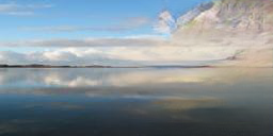}\\
          \vspace{0.1cm}
    \end{minipage}
    }

\vspace{-0.1cm}
\caption{Exhibition of failure cases. Although the model could synthesize contents matching the guiding sketch but fail in predicting reasonable pixels for the desired shape.}
\label{failture-cases}
\vspace{-0.3cm}
\end{center}
\end{figure*}

By including the feature correlations of multiple layers, the multi-scale style representations are obtained, and the total style loss can be calculated accordingly:
\begin{equation}
    \mathcal{L}_s = \sum_{d\in D} w_d\mathcal{L}^d_s,
\end{equation}
where $D$ is the index collection of selected activation layers, and $w_d$ is the trade-off weight. 
In our experiments, the activated output of layer relu\_Y\_1(Y=1,2,3,4,5) of VGG19 network~\cite{vgg19} are taken for style representation, \ie, $|D|=5$. 
The designed style ranking loss is equipped to the generator loss to train the network. 

\section{Experiment}
\subsection{Experiment Setup}
\noindent\textbf{Dataset.}
This work focuses on performing image outpainting for scenery images. We conduct extensive experiments on two benchmarks, \ie, NS6K~\cite{yangzx} and NS8K~\cite{wang}, to validate the effectiveness of our ReGO. The \textbf{NS6K} dataset contains 6040 scenery images in total, and 5040 images are treated as training data while the rest is used for testing~\cite{yangzx}. The \textbf{NS8K}, which consists of 8115 images, contains much more diverse scenery images comparing to NS6K. Of these, 6115 images are taken as training data, the rest is used for testing. The \textbf{SUN Attribute} dataset has 14,340 diverse enough images from 707 scene categories, we randomly select 80\% and 20\% for training and testing, respectively.

\vspace{0.15cm}
\noindent\textbf{Implement Details.}
\label{imple}
\yx{Our proposed ReGO provides a model-agnostic solution and could be plugged into many off-the-shelf outpainting models. In this work, we apply our ReGO to three state-of-the-art outpainting methods, including NSIO~\cite{yangzx}, BDIE~\cite{bds}, and SGIO~\cite{wang}, to validate its superiority:}

\yx{
\noindent \textbf{NSIO~\cite{yangzx}} is originally designed for random content prediction. To achieve the sketch-guided outpainting, we make some modifications for NSIO as follow: the left half sketch is channel-wise concatenated with the input, while the right half sketch is encoded and fed as the initial state of LSTM~\cite{LSTM} decoder to predict the hidden feature of the full images. Our ACS module is plugged after each decoding layer except the last one, and the style ranking loss is weighted by 0.5 and added into the generator loss to train the network. }

\yx{
\noindent \textbf{BDIE~\cite{bds}} is a random-outpainting model as well, and the sketch is concatenated with the input to perform the sketch-guided outpainting. Besides, the conditional skip connection and the position channels in SGIO~\cite{wang} are also equipped to BDIE~\cite{bds}, to build a stronger baseline. The ACS module and the style ranking loss are equipped in an analogous way with NSIO~\cite{yangzx}. }

\yx{
\noindent \textbf{SGIO~\cite{wang}} is the first attempt for the sketch-guided outpainting. The ACS module is also employed after each decoding layer for texture compensating, and the style ranking loss is added to the generator loss with weight 0.5 to ensure the style consistency.}


For our study, the style ranking loss is equipped to the generator loss,  and the weights of style ranking loss in multiple layers are all set as 0.2, \ie, $w_d=0.2$. During the training stage, 
five neighbors are employed in our baseline methods, and the impact of neighbor number will be discussed in the following experiment. At the testing stage, only the most similar neighbor is used to synthesize the outpainting.
Besides outpainting models, we also include three state-of-the-art inpainting models for comparison, \ie \textbf{DeepFillv2}~\cite{Yu0}, \textbf{CoModGAN}~\cite{comodgan}, and \textbf{LaMa}~\cite{lama}. For LaMa and CoModGAN, we mask the right half images and introduce the sketch as an additional channel to train the network, while DeepFillV2 is a sketch-guided inpainting model, and it's trained by only restoring the right half image.  To make a fair comparison, the loss functions, the hyperparameters and the training details all follow the same settings of their original papers. The sketch augmentation strategy~\cite{wang} is also employed for all methods to enhance the free-form outpainting.

\setlength{\tabcolsep}{2.0mm}{
\begin{table}[t]
\caption{The contributions of each part in our method. RI, ACS, and SRL indicate reference image, ACS module, and style ranking loss respectively.}
\vspace{-0.3cm}
\begin{center}
\begin{tabular}{|cccc|ccc|}
\hline
 & RI& ACS& SRL &IS${\color{red} \uparrow}$ &FID${\color{red} \downarrow}$ &MSD${\color{red} \uparrow}$\\
\hline
BDIE~\cite{bds} &  & &  &3.002 & 11.021 &0.963\\
BDIE~\cite{bds} &\checkmark  & &  &2.918 & 10.991 &1.034\\
BDIE~\cite{bds} &\checkmark &\checkmark  & &3.038 &10.269 &1.221\\
BDIE~\cite{bds} &\checkmark &\checkmark &\checkmark  &\textbf{3.126} &\textbf{10.052} &\textbf{1.357}\\
\hline
\end{tabular}
\end{center}
\label{ablation study}
\vspace{-0.2cm}
\end{table}
}

\setlength{\tabcolsep}{2mm}{
\begin{table}[t]
\setlength{\abovecaptionskip}{0pt}%
\setlength{\belowcaptionskip}{2pt}%
\begin{center}
\caption{The performances of the style ranking loss and the style regression loss.}
\label{srl_compare}
\begin{tabular}{|c|ccc|}
\hline
Method &IS${\color{red} \uparrow}$ &FID${\color{red} \downarrow}$ &MSD${\color{red} \uparrow}$\\
\hline
$\text{ReGO}_\text{BDIE}$-Reg & 3.089 & 10.956 &1.163\\
$\text{ReGO}_\text{BDIE}$ &\textbf{3.126} &\textbf{10.052} &\textbf{1.357}\\
\hline
\end{tabular}
\end{center}
\end{table}
}

\subsection{Evaluation Metric}
Following Wang's~\cite{wang} setting, three metrics, \ie, Fr$\acute{\rm e}$chet Inception Distance (FID)~\cite{FID}, the Inception Score (IS)~\cite{IS} and Mean Satisfactory Degree (MSD)~\cite{wang}, are employed for evaluation.
To evaluate the free-form outpainting results, we randomly select 555 images from test data and replace the original sketches with manually drawn free-form ones, 89 different types of sketches are collected in total. 20 volunteers are invited to label the free-form outpainting results as three levels: 0-poor, 1-ordinary and 2-good, and the mean value of all labels on selected images are taken as the mean satisfaction degree (MSD), which is taken for subjective comparison since there is no groundtruth available. Comparing to the FID and the IS, MSD directly reflects the performance in practical situations, therefore, it is a critical metric to evaluate the generalization ability on free-form sketches.

\subsection{Quantitative Comparison}
The performance of both sketch-guided outpainting and random outpainting on datasets NS6K and NS8K is reported in Table~\ref{main_perform}, where $\text{ReGO}_\text{NSIO}$ denotes the NSIO model with our proposed ReGO equipped. From Table~\ref{main_perform}, we can observe that our proposed ReGO module could simultaneously enhance the sketch-guided outpainting and the random outpainting.

\vspace{0.15cm}
\noindent\textbf{Sketch-Guided Outpainting.}
Our proposed ReGO module could boost the performance of three state-of-the-art outpainting and inpainting methods on both NS6K and NS8K. For example, the FID of BDIE~\cite{bds} is 11.021 on NS6K, when our ReGO module is equipped, the FID could reach 10.052. Besides the image restoring according to the original sketches, the free-form outpainting can also be improved by the ReGO module. The MSD of $\text{ReGO}_\text{SGIO}$ can reach 1.201 on NS6K, while the original SGIO's is only 1.01. Our best performance is achieved based on the BDIE~\cite{bds}, which could reach 10.052 FID and 1.357 MSD on NS6K. It is clear from Table~\ref{main_perform} that our proposed ReGO module can both improve the image rebuilding and free-form outpainting on three backbones, which validates the effectiveness of our method. 

\yx{In our ReGO, the searched neighbors of the input serve as the reference, while the input image itself is also an intuitive choice since it naturally shares the highest similarity with the input. To study the effects of two types of references, we conduct experiments with BDIE backbone on NS6K dataset to investigate the performance difference. Let $\text{ReGO}_\text{BDIE}$-SR (\textbf{S}elf-\textbf{R}eference) denotes the $\text{ReGO}_\text{BDIE}$ with the input as reference. In our experiment, $\text{ReGO}_\text{BDIE}$-SR could also achieve acceptable performance on image rebuilding according to the original sketches, however, it shows poor generality when encountering the free-style sketches. For example, the FID of $\text{ReGO}_\text{BDIE}$-SR could reach 10.561 on NS6K and surpasses the method BDIE, however, its MDS for free-form outpainting is only 1.012, which is much worse than $\text{ReGO}_\text{BDIE}$. We guess the barren sketch layout and content pattern cause somewhat overfitting, consequently, the model trained with self-reference could not well generalize to the  free-style outpainting. In contrast, When the neighbors serves as the reference, the model could see diverse training pairs, as a result, the trained model could perform well on both the image rebuilding and free-form outpainting.}

\vspace{0.15cm}
\noindent\textbf{Random Outpainting.}
Besides providing the sketches to harvest the desired outpainting, another possible scenario is that the users may refuse to drawn any guiding sketches and only attempt to obtain the random results. How would our system perform if no guiding sketches are fed? To validate the effectiveness of our method under such a scenario, we conduct experiments to predict random results and report the performance on both datasets. Since the NSIO~\cite{yangzx} and BDIE~\cite{bds} are originally designed for random outpainting, we follow the same pipelines as their original papers~\cite{yangzx,bds} to train the networks. As for the inpainting methods,  CoModGAN~\cite{comodgan}, and LaMa~\cite{lama}, we directly mask the right half of the image for training. For sketch-guided systems, we simply set the right half sketch as zeros to conduct random outpainting.

The results are also reported in Table~\ref{main_perform}, we can observe that abandoning the original guiding sketches significantly damnify the performance of the sketch-guided outpainting systems. For example, the $\text{ReGO}_\text{SGIO}$ with guiding sketches could reach 10.104 FID on NS6K, while its FID w/o the guiding sketches deteriorates to 15.396, this is because the systems are trained with the original sketches. The performance of SGIO with our ReGO module is comparable with its original method SGIO~\cite{wang}. For the random prediction methods, $\text{ReGO}_\text{NSIO}$ performs slightly worse than the NSIO~\cite{yangzx}, while $\text{ReGO}_\text{BDIE}$ is more outstanding comparing to the STOA outpainting (BDIE~\cite{bds}) and inpainting methods (LaMa~\cite{lama}). From Table~\ref{main_perform}, we can see that even though no guiding sketches are provided, the methods with our ReGO module could also produce comparable results with the original methods. With the designed ACS module, we can develop an unified framework that could simultaneously deal with the random prediction and the sketch-guided outpainting. what's more, the BDIE with ReGO module could achieve the SOTA performance on both tasks.

\subsection{Ablation Study}

\noindent\textbf{Validate the Components of ReGO.}
To validate the contributions of each component in our proposed ReGO, we employ the BDIE~\cite{bds} as backbone and conduct ablations on the NS6K dataset.  
Quantitative results are reported in Table~\ref{ablation study}. 


\yx{As shown in Table~\ref{ablation study}, when the reference image is introduced, the FID of BDIE could be improved from 11.02 to 10.99 and the MSD could also be boosted, which reveals compensating the texture details from the neighbors is a promising idea. However, only the reference image does not make the performance outstanding enough. With the proposed ACS module adopted, the network could adaptively block the profitless content and distill the beneficial pixels, and we can observe a significant performance improvement where the MSD and the FID could reach 1.221 and 10.269, respectively.  When the style ranking loss is further equipped to ensure the style consistency, the performance step further. The model with three parts simultaneously utilized achieves the best performance, and when a new mechanism is equipped, the performance gets improved, which validates the contributions of each component.}

Fig.~\ref{ablation_rebuilding_and_free_form} exhibits the visual results of the ablation comparison on image rebuilding and free-form outpainting. From Fig.~\ref{ablation_rebuilding_and_free_form}, the contribution of each component can be clearly observed.

\vspace{0.1cm}
\noindent\textbf{Validate the Style Ranking Loss.}
To produce the style-consistent results, we design the style ranking loss to prevent the style of the synthesized content being affected by the reference image. Intuitively, the close style between the synthesized content and the input can also be achieved by directly conducting a regression procedure. In this subsection, we study the impact of these two solutions.

Table~\ref{srl_compare} shows the performance under IS, FID, and MSD, where $\text{ReGO}_\text{BDIE}$-Reg indicates $\text{ReGO}_\text{BDIE}$ using the $l_2$ style reconstruction loss instead of the style ranking loss. From Table~\ref{srl_compare}, the $\text{ReGO}_\text{BDIE}$ with the proposed style ranking loss performs much better on both image restoring and free-form outpainting.  The reasons for such results may stem from the overfitting. Although the left half and the right half parts are from the same image, the style representations captured by the Gram matrices are still different. Therefore, stiffly conducting such a regression procedure is easy to cause overfitting. Besides, the main goal of our style ranking loss is to block the style of the reference image being reflected on the extended content, the style consistency between the input and the synthesized part is not necessary to enhance, since it can be achieved by the pixel-wise reconstruction and the adversarial training~\cite{bds,yangzx,wang}.

\vspace{0.2cm}
\noindent\textbf{Discuss the Number of  Reference Image.}
In our baseline methods, five neighbors for each training sample are selected, and we randomly pick up one in each iteration to serve as the reference image. This subsection investigates the impacts of the number of the reference image.

The performance tendencies with five different reference numbers are shown in Fig.~\ref{dis_refnum}, where the FID is scaled by the logarithmic function.  As shown in Fig.~\ref{dis_refnum}, there are two important observations. First, Comparing to employing only one reference image, using multiple references could enhance the model generality and train a more robust generation model. The FID of $\text{ReGO}_\text{BDIE}$ with only one reference image is 10.728, when the reference number increases to 5, the FID could be improved to 10.052. Second, more reference images do not make the performance step further, which is clear from the performance tendencies when the reference number ranges from 5 to 20.  The situation with five reference images achieves the best performance on average.

\subsection{Qualitative Results}
\noindent\textbf{Image Rebuilding.}
Fig.~\ref{rebuiling_and_RD_outpainting} provides the visualizations of the rebuilding results according to the original sketches and random outpainting for SOTA inpainting and outpainting models. \yx{To ease the visual exhibition as well as saving some space, we only exhibit the outpainting results of our best model $\text{ReGO}_\text{BDIE}$. It can be observed that the results of $\text{ReGO}_\text{BDIE}$ are more authentic and natural due to the richer textural details.} 

From Fig.~\ref{rebuiling_and_RD_outpainting} I, the comparison methods, LaMa~\cite{lama}, SGIO~\cite{wang} and BDIE~\cite{bds}, could extend reasonable pixels for the input image, but the predicted content is blurry and lacks textural details, which makes the overall image not authentic enough. While $\text{ReGO}_\text{BDIE}$ could produce texture-rich outpainting results. The results of random prediction are exhibited in Fig.~\ref{rebuiling_and_RD_outpainting} II, comparing to the competing methods, $\text{ReGO}_\text{BDIE}$ could also successfully synthesize the results with more textural details when no sketches are fed, and the synthetic images are even more satisfactory than the method original designed for the random prediction, \ie, BDIE~\cite{bds}. From Fig.~\ref{rebuiling_and_RD_outpainting},  we could find that the guiding sketch is not one of requisite inputs for our system, when the users do not provide the guiding sketch, ours system could also produce satisfactory random outpainitng results. 

\vspace{0.15cm}
\noindent\textbf{Free-form Outpainting.}
The comparison for free-form outpainting is exhibited in Fig.~\ref{free_form comparison_transfer_guide} I, $\text{ReGO}_\text{BDIE}$ could not only synthesize the expected content matching the guiding sketch but achieve authentic and natural enough results. Especially the boundaries of different semantic regions are much clearer than the competing methods. Additionally, we surprisingly find that the reference image could also help fill reasonable pixels for the free-form outpainting, as shown in the bottom row in Fig.~\ref{free_form comparison_transfer_guide} I. Besides the manually drawn sketches, we could also use the sketch from another image to guide the outpainting, as shown in Fig.~\ref{free_form comparison_transfer_guide} II. The inputs in Fig.~\ref{free_form comparison_transfer_guide} II(a) directly use the sketches of the reference images to control the outpainting, while the ones in Fig.~\ref{free_form comparison_transfer_guide} II(c) use the sketches from randomly selected images, two types of cases serve as the simple and the difficult cases, respectively. From Fig.~\ref{free_form comparison_transfer_guide} II, our method could not only predict new content matching the guiding sketches but achieve satisfactory style-consistency for both simple and difficult cases. It's worth noting that this paper mainly focuses on synthesizing new content along left to right, however, the prediction of other directions could also be performed based on BDIE backbone, just as shown in Fig.~\ref{full_direction}, we leave everything unchanged except for using the mask to indicate the missing regions.  In this task, our $\text{ReGO}_\text{BDIE}$ could also achieve more outstanding performance comparing to the BDIE model, 10.817 (FID) 3.004 (IS)-$\text{ReGO}_\text{BDIE}$  VS 12.132 (FID), 2.893 (IS)-BDIE.

\vspace{0.15cm}
\noindent\textbf{Results on High-Resolution Images.}
Besides the low resolution dataset, we also collect 558 high-resolution scenery images from Internet using the key word "scenery images" to further evaluate our model. We resize the images as 512$\times$768  and directly evaluate the performance on this dataset, the performance of our method could also outperform the most competitive method BDIE, 34.012(FID) 4.078(IS)-$\text{ReGO}_\text{BDIE}$ VS 37.841(FID) 3.917(IS)-BDIE. Fig.~\ref{highreso} shows the high resolution results of three state-of-the-art methods.

From the above, the proposed framework allows users to harvest three types of results: random outpainting, free-form outpainting from manually drawn sketches and controllable outpainting using sketch from another image. Therefore, our proposed method is with higher practical value. We also show some failure cases in Fig.~\ref{failture-cases}, the model fails to predict reasonable pixels or generate natural enough contents for the guiding sketches. In our experiment, we find most of failed samples are from those whose guiding shapes are in the top part of the images, we guess the reason stems from the barren sketch patterns of training images. This reveals that there are still many challenges standing for this task. In the future, we will continue to explore this task and attempt to address more issues of this field.

\section{Conclusion and Future Work}
This work develops a novel ReGO module that improves the outpainting quality by borrowing pixels from its neighbors. The proposed method could  effectively boost the results of the sketch-guided image outpainting by enriching the textual details. An ACS module is proposed to distill the beneficial pixels and suppress the profitless contents, which could help the generator pick up the helpful pixels. To prevent the style of the synthesized content being affected by the reference image, we design a style ranking loss to enforce the generator to produce the style-consistent contents. Experiments on two benchmarks based on three backbones demonstrate the effectiveness of the proposed method. The idea that enriches the details from neighbors may also work for other generation tasks, we will continue to explore the effectiveness of the proposed method in the future.

\end{document}